\documentclass[a4paper,fleqn]{cas-sc}
\usepackage[numbers,sort&compress]{natbib}
\usepackage{amsmath,amssymb}
\usepackage{amsthm}
\usepackage{booktabs}
\usepackage{graphicx}
\usepackage{multirow}
\usepackage{pifont}
\usepackage{float}
\usepackage{colortbl}
\definecolor{okgreen}{RGB}{0,110,60}
\definecolor{badred}{RGB}{190,35,35}
\definecolor{hlcell}{RGB}{219,233,247}
\providecommand{\cmark}{\ding{51}}

\newtheorem{theorem}{Theorem}

\newtheorem{remark}{Remark}

\begin{document}

\renewcommand{\topfraction}{0.85}
\renewcommand{\bottomfraction}{0.5}
\renewcommand{\textfraction}{0.1}
\renewcommand{\floatpagefraction}{0.75}

\shorttitle{When Rankings Move}
\shortauthors{Lu et al.}
\title[mode=title]{When Model Rankings Move Without Model Change:
A Registered Audit of Evaluation Contracts in Neural Prediction Pipelines}

\author[aeu]{Hao Lu}[orcid=0009-0005-3338-8800]\cormark[1]
\credit{Conceptualization, Methodology, Software, Investigation, Data curation, Writing -- original draft, Writing -- review \& editing}
\ead{luhao_0210@163.com}
\author[aeu]{Weixing Zhu}
\credit{Supervision, Writing -- review \& editing, Project administration}
\author[aeu]{Ruiyang Li}
\credit{Data curation, Verification, Writing -- review \& editing}
\cortext[1]{Corresponding author}
\affiliation[aeu]{organization={Army Engineering University of PLA},
city={Nanjing}, postcode={210007}, country={China}}

\begin{abstract}
Evaluation reports cite sliced differences,
  most prominently the intra--inter gap in document-level relation
  extraction (DocRE), as capability evidence.  Such gap readings are
  joint properties of models and \emph{evaluation contracts}: scoring
  function, gold version, threshold, output format, and
  prompt language.  With predictions frozen, switching only the gold
  reverses a method ordering ($+0.0363$, CI excluding zero; retrained
  stand-ins) , leaving it tie-level.  The flip survives Holm
  correction (exploratory tier); ten of thirteen flips share one contrast
  (fixed vs decoded threshold).  Rename-only revisions provably cannot move a reading
  ($\Delta = 0$); re-annotation moves it $+0.862$ UD points.
  We formalize this as an \emph{evaluation leverage account}
  centered on a rank-flip condition.  Constructively, the
  \emph{audited evaluation reading} (AER; post-freeze) aggregates a
  comparison over declared contracts, declares ties when they
  disagree,
  prescribes seeds, and filters
  unverifiable settings.  On WMT19
  it is indistinguishable from the best single contract at the
  exploratory, convention-stable margin $\epsilon{=}0.10$ (chosen
  post-freeze); its
  tie declarations cover every contract-induced flip (14/14),
  at a priced false-tie cost (66.1\%/88.0\% of declared ties; single
  contracts: 58.7--71\% de-en, 86--91\% zh-en).
  Coverage drops to 73\% held out and 3/13 on the denser DocRE
  registry (declare-tie-if-any-flip: 89\%): the guarantee is
  grid-specific.  Evidence spans DocRE, HacRED, a
  UD probe, released gold revisions, and a replication.  Failed
  preregistered predictions are retained as boundaries.  A 5\%
  violation gate bounds survivor-selection bias.  We release
  EvalLever, a registered audit workflow (147 amendments, tiered status).  On a three-judge WMT19 grid its
  ensemble suppresses part of the sign instability (13--15 to
  9; one
  preregistered criterion fails)
\end{abstract}

\maketitle

\section{Introduction}
\label{sec:intro}

Evaluation reports steer NLP and, with language models now embedded in
larger intelligent systems, increasingly steer deployed-AI decisions
(model selection, A/B comparisons, release regression) outside it; yet
trust in those readings is
declining: rankings disagree across benchmarks, results fail to
replicate under seed resampling, and selective reporting skews the
literature.  Though treated separately, these ailments share a root cause:
for a widely used family of readings (sliced performance
differences, what we call the sliced-gap readings), each reading is a
function of \emph{evaluation contracts} that are rarely frozen,
reported, or audited.

Each mechanism has been observed piecemeal, but treated as unrelated anomalies: gold revisions reorder rankings \citep{tan-etal-2022-revisiting}, metric reuse distorts readouts \citep{pombal-etal-2025-mint}, prompt formatting shifts scores \citep{sclar-etal-2024-quantifying}, label-error audits shift reported performance \citep{nahum-etal-2024-label,neuhaus-hein-2025-repop}. To our knowledge, such choices have not been formalized as a contract object with measurable coordinates, a derived invariance, a registered flip condition, or tie-aware readings covering contract-induced flips; these are the steps this paper takes, while frozen-prediction re-scoring under released golds has precedents \citep{northcutt-etal-2021-pervasive,gema-etal-2025-mmlu,alt-etal-2020-re-tacred} that we extend rather than replace.
Propositions~2--3 formalize the rank-flip condition and the variance ordering; the empirical sections test their assumptions, and Sections~\ref{sec:audit}--\ref{sec:paradigm} turn the account into an audit protocol.  Section~\ref{sec:nn-implications} distills the architecture-level findings for the neural-network designer.

\begin{figure}[htbp]
\centering
\includegraphics[width=\columnwidth]{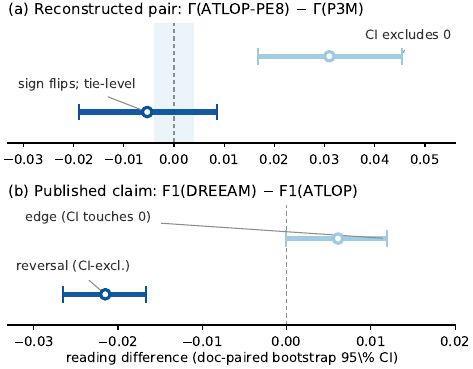}
\caption{One claim, two readings: the registry's headline gold-swap flip (ATLOP-PE8$|$P3M) under the original (gD) and revised (gR) released golds, with doc-paired bootstrap 95\% CIs; the ordering reverses and lands at tie level under gR. Full flip detail: Figure~\ref{fig:flipsforest} and Section~\ref{sec:empirical-flip}.}
\label{fig:argflip}
\end{figure}

Consider two published methods for document-level relation extraction
(DocRE)~\citep{yao-etal-2019-docred} on the same 498-document DocRED
development subset (189{,}556 frozen candidate pairs): an ATLOP-style
encoder~\citep{zhou-etal-2021-atlop} and the PU-metric-learning model
P3M, audited through retrained instantiations (released checkpoints
run $-3.6$ to $-10$ points below their published numbers through their
own entrypoints; Appendix~E (supplementary)).
Under the original DocRED gold standard (gD), the intra--inter gap
of the ATLOP-style variant exceeds P3M's by $+0.0309$ (CI
$[+0.0167, +0.0455]$, excluding zero).  Switch only the gold
standard, to the revised Re-DocRED projection (gR), published
precisely to fix false negatives, and the ordering \emph{reverses}
($-0.0053$, CI $[-0.0190, +0.0086]$); the flip magnitude $+0.0363$
(CI $[+0.0204, +0.0529]$) excludes zero: the registry's gold-axis
flip, exploratory tier but Holm-surviving (\S\ref{sec:empirical-flip}).  Neither model changed a single prediction; the
arm-level movement replicates on a held-out revised population for the
post-review levers with retained checkpoints (\S\ref{sec:empirical-llm}).  A claim of the form ``our model narrows the intra--inter gap'' is thus not guaranteed to be a property of the models: it is jointly a property of the models and a contract the claim never states.  Such claims are routine: the intra/inter split dates to the DocRED annotation itself \citep{yao-etal-2019-docred}, with tabulated intra/inter F1 tracing to the latent-structure line and persisting as a reporting convention since the early DocRE era \citep{nan-etal-2020-lsr,sahu-etal-2019-intersent,xu-etal-2021-ssan}, and recent DocRE systems report the intra--inter gap as a headline evaluation \citep{zhang-etal-2025-global}; the gap-as-capability reading has design precedent in separate intra-/inter-sentential reasoning \citep{zeng-etal-2021-sire}; and the gap's false-negative mass motivates method design itself \citep{wang-etal-2024b-p3m}.  Figure~\ref{fig:argflip} previews both readings with paired CIs: the same comparison reads as an edge, a tie, or a reversal depending on the contract alone.  The phenomenon is axis-concentrated: across the full registry, ten of thirteen flips sit on the threshold axis (a policy knob some systems optimize internally), the gold axis carries the one further CI-excluding reversal above plus two CI-including flips, and the scoring axis contributes none (\S\ref{sec:empirical-flip}); structured prediction is the densest testbed, and \S\ref{sec:cross-task} carries the account across task families beyond it.

\looseness=-3
We contribute (i) a \emph{formal account of evaluation leverage} (four propositions with explicit assumptions and tested boundaries, all projections of one reading $\rho(M,c)$; the lemmas are elementary (permutation symmetry, triangle inequality, variance additivity) and are claimed for the discipline they impose, not depth), stating when contract changes move readings and rankings, and when they provably cannot: rename-only revisions are gauge transformations whose \emph{arm} is zero by derivation (Section~\ref{sec:theory}); (ii) a \emph{demonstration across task families}: with predictions frozen, a reconstructed ordering reverses under the gold standard at the point-estimate level, arms recur across the remaining coordinates (flips only where arms are long and signs contrastive), with a violation surface for prompted predictors; negative results retained (Sections~\ref{sec:empirical}--\ref{sec:audit}); (iii) a \emph{construction}: the \emph{audited evaluation reading} (AER: contract marginalization, tie declarations, a measurement prescription, and a validity gate), validated against official human anchors (themselves contract coordinates, \S\ref{sec:aer}), matching the best single metric contract at an exploratory $\epsilon{=}0.10$ equivalence on the WMT19 grids (chosen post-freeze; a point-level edge on de-en only), tie declarations covering every flip on the declared grids (73\% held out; 3/13 on the denser DocRE flip registry; the guarantee is grid-specific; Section~\ref{sec:aer}); and (iv) a \emph{prescription}: freeze, state, and audit the contract, with seed prescriptions, evidence tiers, and EvalLever\footnote{The complete evidence bundle accompanies this submission as supplementary material (a sha256-manifested archive: code, frozen per-document predictions for all nine task families, the contract registry, id-level multi-gauge readings, the preregistration trail, 147 amendments, each timestamped, with an English one-line summary sheet, and every cited statistics file); it is public at submission (\url{https://github.com/lutttw/evallever}, v1.0.2, tag-pinned; the provenance-attestation scope is stated in the Data availability section; snapshot DOI 10.5281/zenodo.24442304).  Identifiers in the text refer to files and result families in this bundle: r1/r26 = flip registry and Holm pass, r2 = seed convergence, r24/r36--r47 = boundary and mechanism probes, r49--r52 = cross-task replications, r54--r58 = judge grid, claims audit, and TOST null bounds, r59--r61 = AER construction and validation, r64/E-A* = post-review analysis extensions, r64/E-C* = judge-grid and coverage repairs, N1--N4 = released-gold audits; in-text codes: C1--C3 = confirmatory comparisons, CL1--CL3 = published-claim rows, H-G1/H-X1/H-J3/J4 = registered hypotheses, B1--B5 = boundary conditions, V1--V4 = random-source levels.} (Sections~\ref{sec:audit}--\ref{sec:paradigm}).  Together they form a normative baseline (freeze, report with ties, gate) at checklist cost (Section~\ref{sec:paradigm}) on any task satisfying the applicability criterion.  Where the data contradict a preregistered prediction we say so. Scope: deeply validated on DocRE, HacRED, and a UD probe; boundary results (Section~\ref{sec:boundary}) mark where it fails.

\noindent\fbox{\parbox{\dimexpr\columnwidth-2\fboxsep\relax}{\small
\textbf{The protocol in one paragraph.}  For a practitioner the paper's prescription is: \textbf{freeze} the contract (state scorer, gold version, threshold, format, language before comparing); \textbf{audit} by re-scoring frozen predictions under the declared alternatives (the \texttt{evallever} stack: \texttt{register} the contract set, \texttt{aer} reads the marginal reading with ties declared, \texttt{audit} scans violations, \texttt{seed} prescribes a contrast-specific minimum seed count); and \textbf{report} edges, ties, and reversals as contract-indexed readings rather than single numbers.  Sections~\ref{sec:audit}--\ref{sec:paradigm} develop each step; the whole pass runs on CPU over frozen predictions (audit layer; prediction generation is separate).
}\medskip}

\section{Related Work}
\label{sec:related}

Each coordinate of the evaluation contract has its own literature; this
section situates the account among them.

\paragraph{Gold revision and evaluation bias.}
\looseness=-1
Gold revisions reorder rankings: Re-DocRED \citep{tan-etal-2022-revisiting}, TACREV/Re-TACRED \citep{alt-etal-2020-re-tacred,stoica-etal-2021-retacred}, label-error re-annotation \citep{nahum-etal-2024-label,neuhaus-hein-2025-repop,gema-etal-2025-mmlu}, the DocRED recommend--revise pipeline's own missing-instance audit \citep{huang-etal-2022-recommend}, documented prevalence of false-negative golds \citep[post-hoc error analysis]{xie-etal-2022-eider}, re-scoring under corrected labels \citep[on the correctable subset]{northcutt-etal-2021-pervasive}, and, on the training side, interventional document revisions followed by retraining \citep{nadkarni-2026-rewriting}.  ReCOGS's rename-invariance \citep{wu-etal-2023-recogs} and guideline-centered annotation \citep{ruggeri-etal-2026-guidelines} presage our gauge principle; rename-invariance checks are also a standing instrument in software-engineering testing, where the gauge is an instance of a metamorphic relation \citep{segura2016survey}.  We complement: holding predictions fixed, we measure the leverage of switching among released golds, rather than producing a corrected dataset or retraining.  Closest in spirit, Opitz~\citep{opitz-2024-closer} quantifies rank movement across classification metrics and Chen and Tanaka-Ishii~\citep{chen-tanaka-2026-benchmark} diagnose under-specification in benchmark language interpretation; neither holds predictions fixed while switching among released golds, and neither derives when such switches provably cannot move an ordering.  In ML meta-science, benchmark validity has a parallel philosophical line: construct-validity frameworks for model evaluation \citep{freiesleben2026benchmarking} and critiques of overgeneralized benchmark claims \citep{raji2021ai}; this audit is complementary, asking not what a benchmark measures but when a fixed benchmark's own contract coordinates move the reading.  Concurrent work likewise revisits generalizability theory for AI-era assessment \citep{degen2025revisiting}; our mapping differs in object (evaluation contracts for model comparison rather than rater reliability) while sharing the facet logic.

\paragraph{Variance, seeds, and significance.}
\looseness=-1
Variance protocols address training-side variance \citep{bouthillier2021accounting}; significance testing and power audits address the evaluation side \citep{dror-etal-2018-hitchhiker,card-etal-2020-little}; conformal evaluation attaches per-reading coverage \citep{zerva-martins-2024-conformalizing}.  We find variance structured by the randomness level of its effect path (Section~\ref{sec:empirical-llm}).

\paragraph{Rank stability.}
\looseness=-1
Rank reversal under benchmark perturbation, prompt formatting, and answer extraction is the empirical backdrop \citep{bouthillier2021accounting,dodge-etal-2019-show,mizrahi-etal-2024-state,sclar-etal-2024-quantifying}; item-level analyses go one grain finer, showing that leaderboard examples are not equally informative and that aggregate scores can mask this \citep{rodriguez-etal-2021-evaluation}.  Evaluation-set composition sweeps exist as well: Reyes-Montesinos et al.~\citep{reyes-montesinos-etal-2025-none} cross an $11{\times}11$ train--test grid over unanswerable ratios and conclude that single-scenario scores are over-optimistic, but each cell retrains a fresh model, the sweep covers one dataset-level contract, and the takeaway is a design guideline rather than an audit.  Closest to our constructive layer, Ceron et al.~\citep{ceron-etal-2024-beyond} gate readouts on a battery of reliability tests and already treat near-zero outcomes as first-class. AER differs in composition: it registers and marginalizes over contracts up front (ties and prescribed seeds) rather than gating first and aggregating survivors.

\paragraph{LLM-as-judge and prompted evaluators.}
\looseness=-1
LLM judges correlate with humans \citep{zheng-etal-2023-mtbench,liu-etal-2023-geval,verga-etal-2024-poll} while positional and self-preference biases persist \citep{wataoka-etal-2024-selfpref,wang-etal-2024-fair}; inter- and intra-judge variance have been probed under ambiguity-inducing prompts and across rating runs \citep{kumar-2026-probing,stureborg-etal-2024-inconsistent}; Zheng et al.~\citep{zheng-etal-2023-mtbench} anticipate cross-judge score instability qualitatively, yet none of these audits the judge as one coordinate among the contract set our framework treats jointly (identity, rubric, format, language, gold version, threshold), with tie marginalization (\S\ref{sec:aer}).  Concurrent work \citep{chang-etal-2026-padme,pakhomov-nijkamp-2026-frozen} and Gilda and Gilda~\citep{gilda-gilda-2026-perishable} likewise call scores perishable; we operationalize the apparatus: contract object, invariance, frozen-prediction audits, constructive reading.

\paragraph{Metric-choice sensitivity and this account.}
\looseness=-3
Metric-choice sensitivity extends beyond NLP \citep{musgrave-etal-2020-metric,deutsch-etal-2021-statistical,marie-etal-2021-scientific,kobayashi-etal-2024-revisiting,stengel-eskin-van-durme-2023-calibrated,jiang-etal-2024-rethinking,mathur-etal-2020-tangled,freitag-etal-2021-experts}; the WMT metrics tasks evaluate evaluators annually \citep{ma-etal-2019-metricstask,freitag-etal-2023-metricstask,polak-etal-2025-latency}.  Each examines \emph{one} instrument facet and quantifies rank movement within it; none combines frozen predictions, \emph{officially released} alternative contracts, arms, registered flip checks, and a derived invariance in one audit (Table~\ref{tab:delta}).  Benchmark Agreement Testing compares model rankings across different benchmarks to validate candidate suites \citep{perlitz-2024-bat}, it asks whether a new benchmark reproduces the established one; the present audit holds the benchmark fixed and shows the scoring contract within a single benchmark moves rankings by itself.  Checklist-style behavioral testing \citep{ribeiro-etal-2020-checklist} and holistic multi-scenario evaluation \citep{liang-etal-2024-helm} widen what is audited (many scenarios under one harness) but the harness's own contract coordinates remain fixed, so contract movement is not the measured quantity.  Nor is the gap hypothetical in the benchmarking genre itself: recent studies declare a single winner across eight datasets under one frozen F1@K scoring contract while switching the metric across document-length regimes \citep{martinez-cruz-etal-2024-chatgpt}, contract sensitivity of exactly the kind Section~\ref{sec:empirical-flip} quantifies.  Recent trustworthy-AI surveys likewise flag, at taxonomy level, that reported results across studies ``cannot be compared directly'' \citep{xue-etal-2026-trustworthy}; what is missing is precisely the contract object of Section~\ref{sec:theory}.

\begin{table}[t]
\centering
\caption{What each evaluation-audit line of work leaves unmeasured.}
\label{tab:delta}
\setlength{\tabcolsep}{3.2pt}
{\scriptsize
\begin{tabular}{@{}lcccc@{}}
\toprule
\textbf{Line of work} & \textbf{Pred.} & \textbf{Legal} & \textbf{Der.} & \textbf{Tie} \\
 & \textbf{frozen} & \textbf{gold} & \textbf{inv.} & \textbf{aware} \\
\midrule
Metrics tasks (2023) & \cmark & \ding{55} & \ding{55} & \ding{55} \\
Label-error re-scoring (2021) & \cmark & \ding{55} & \ding{55} & \ding{55} \\
Gold re-scoring (2025) & \cmark & \cmark & \ding{55} & \ding{55} \\
Gold revision: TACREV (2020) & \cmark & \cmark & \ding{55} & \ding{55} \\
Gold revision: Re-TACRED (2021) & \ding{55} & \cmark & \ding{55} & \ding{55} \\
Input perturbation (2024) & \ding{55} & \ding{55} & \ding{55} & \ding{55} \\
Metric resampling (2021) & \cmark & \ding{55} & \ding{55} & \ding{55} \\
\midrule
This paper & \cmark & \cmark & \cmark & \cmark \\
\bottomrule
\end{tabular}
\par\vspace{3pt}\footnotesize
\emph{Notes:} pred.\ frozen = scoring over fixed outputs; legal gold = switching
among officially released gold-side alternatives only (scorer-side and
format-side alternatives are tracked under Pred.\ frozen; a gold revision qualifies
when it is a human-expert re-annotation under a written protocol,
officially released. TACREV does. Re-TACRED additionally revises the
guideline (41$\to$40 inventory), counted as legal with that caveat;
model-in-the-loop relabelings and ad-hoc
corrections do not, \S\ref{sec:theory}); derived inv.\ = a proven
invariance separating normative from substantive changes; tie aware = ties as first-class ranking output.  Pred.\ frozen covers $s,g,\tau$; $f,\ell$ regenerate outputs by construction (Section~\ref{sec:formal}).  Row justification: TACREV's headline numbers re-score its \emph{released} per-model predictions under the revised gold (frozen outputs; retrained baselines reported separately there). Re-TACRED's headline numbers \emph{retrain} on the revised corpus, its frozen-prediction cross-evaluations appear only as secondary cells, so only TACREV is marked frozen, and our Section~\ref{sec:cross-task} audit re-scores the released TACREV predictions under both revised golds.  The closest constructive comparator (\citealp{ceron-etal-2024-beyond}) gates and then aggregates across scenario sets; it neither derives invariances nor marginalizes over a declared contract set (\S\ref{sec:related}), so it has no row here rather than a partially-matching one.
}

\end{table}

\section{An Evaluation Leverage Account}
\label{sec:theory}

The gaps catalogued above call for a formal object; this section supplies one.  (notation summary: Appendix Table~S.6 (supplementary).)

The four propositions are projections of one reading
$\rho(M,c)$: Proposition~1 fixes $c$ and varies $M$;
Proposition~2 evaluates the same reading at two contracts and asks
when model orderings survive. Proposition~3 decomposes the variance
of $\rho$ across randomization levels. Proposition~4 separates its
geometry from output validity.  Proposition~2, with the rename
invariance of Section~\ref{sec:formal}, is the account's center.
Status: the rename invariance (\S\ref{sec:formal}) is derived, and
Propositions~1--3 each carry a derived formal core, Lemma~A (the
disagreement-set upper bound), Lemma~B (the no-flip margin
guarantee), and Theorem~C (the variance ordering under nested
randomization), whose assumptions and residual slack the empirical
sections test. Proposition~4 remains observational.

\subsection{Formalization}
\label{sec:formal}

\looseness=-3
An \textbf{evaluation contract} $c = (s, g, \tau, f, \ell)$ specifies a scoring function $s$, a gold-standard version $g$ (officially released alternates, not ad-hoc corrections), a decision-threshold policy $\tau$, an output format $f$, and a prompt language $\ell$, in metrology's terms, the object collects what the VIM calls influence quantities \citep{jcgm2012vim}, with the difference that our influence quantities are chosen by the evaluator.  In generalizability-theory terms \citep{cronbach1972dependability,brennan-2001-gtheory}: the contract coordinates are facets of measurement; a single-coordinate switch is a one-facet manipulation; the arm is a facet main effect; a flip is the facet$\times$object interaction changing sign. AER's marginalization averages over the declared facet set, treated as fixed rather than random facets, so generalization is bounded by the declared set (\S\ref{sec:aer}); and the seed prescription $n^{\ast}$ plays the D-study role.  What is not standard in that vocabulary: the evaluator chooses the facets, the legal facet levels are officially released artifacts rather than design choices, and ties are first-class readings rather than failures to reject. The \textbf{evaluation reading} $\rho(M, c)$ is any real-valued quantity an evaluator reports as capability evidence; the \textbf{sliced gap} $\Gamma(M, c) = s(\text{intra} \mid M, c) - s(\text{inter} \mid M, c)$ (neither a property of the model alone nor of the contract alone, but of their interaction) is the primary instance studied.  Propositions are stated on $\rho$; empirics instantiate it at $\Gamma$ (other instances in \S\ref{sec:cross-task}) unless noted.
Throughout: \emph{arm} $= |\Delta\rho|$ (movement under a contract change); \emph{flip} (rank reversal); \emph{violation} (output failing structural grounding; notation: Table~S.6 (supplementary). Predictions are fixed for $s,g,\tau$; $f,\ell$ regenerate outputs; the realizing parser is a de-facto sixth axis folded into $f$ and audited.  \emph{Fixed coordinates ($\kappa$).} Beyond $c$, readings depend on fixed coordinates $\kappa$ (aggregation level, pool composition, anchor set, evaluated population); the intra/inter grouping rule rides on $s$, declared switchable (H-G1, Table~S.7 (supplementary)).  All experiments freeze $\kappa$ and disclose it with the registry; the axis list is not claimed exhaustive.  \emph{Operation levels.} A \textbf{contract switch} replacing one contract with another is \textbf{axis-aligned} when the two differ in exactly one coordinate; the \textbf{legal contract pair} $(c_1,c_2)$ is the object over which arms and the flip condition (\S\ref{sec:prop2}) are quantified; all switches below are axis-aligned unless stated.

\emph{Rename invariance (gauge).}  \emph{Statement.}  A gold revision
that only renames the label inventory, $g' = r \circ g$ for a
bijection $r$ on labels, applied by convention to both sides of every
pair (gold and predictions relabeled by the same $r$; the conjugate
action), cannot move a reading whose scorer factors through the joint (prediction,~gold) partition symmetrically (accuracy and full-inventory micro-/macro-F1 do; subset-restricted variants that exclude a distinguished class are name-anchored and satisfy the premise only when $r$ preserves the inventory).
\emph{Proof.}  The rename applies one common label permutation to both indices of the joint-partition cell counts, which symmetric statistics cannot see; hence $\rho(M,(s,g',\tau,\dots)) = \rho(M,(s,g,\tau,\dots))$ exactly (full proof: Appendix~B (supplementary)).  A gold-side-only relabeling (predictions still matched by their unchanged string identity) would move accuracy; the synthetic check below returns $\Delta = 0.000000$ under the implemented convention, confirming the double-sided application.
\emph{Consequences.} Only re-partitioning revisions can move a reading; scorers referencing label identity outside the gauge (name-anchored class weights, soft string matching) do not factor through the partition \emph{invariantly} and can move under renames alone; psychometrics' counterpart is facet-level invariance (generalizability theory \citealp{cronbach1972dependability,brennan-2001-gtheory}); the cross-group measurement-invariance literature \citep{meredith-1993-invariance} addresses populations rather than renames. A synthetic six-class check confirms both halves (accuracy/micro-/macro-F1 $0.000000$ exactly under a pure rename; $-0.003$ for a name-anchored weight vector), as do the UD slices (renames $\Delta = 0.000$, re-annotation $+0.862$) and the exact character/word null (B2, Section~\ref{sec:boundary}).

Three quantities share the name ``arm''.
The first is the \textbf{error-set agreement} (Jaccard similarity of error sets, a between-model quantity).
The second is the \textbf{disagreement-referenced leverage} $\mathrm{arm}_{\mathcal{D}} = |\hat{Y}\cap\mathcal{D}|/|\mathcal{D}|$ over the two golds' disagreement set (model-vs-gold; the two need not agree, B1), and the third the \textbf{realized shift} $|\Delta\rho|$, an ex-post outcome.  Henceforth an unqualified ``arm'' denotes $|\Delta\rho|$ (Table~S.9 (supplementary)).

The instruments need two testbed properties: \emph{legal contract pairs} (same-data changes any evaluator could equally have chosen) and a dense \emph{disagreement set} between alternatives; structured prediction supplies both, single-gold classification neither (B1, Section~\ref{sec:boundary}). The applicability criterion (sufficiency tested, necessity not): retrieval and MT satisfy both; code generation the contract-pair criterion.

\subsection{Proposition 1: Generator-Dependent Leverage}
\label{sec:prop1}

\emph{Statement.} With data and the switched axis fixed, only the predictor varying, the realized shift $|\Delta\rho|$ grows with prediction--disagreement overlap; the claim is made at \emph{family} resolution (fine-tuned encoders vs.\ prompted models, plus two within-family slices), with no monotone curve claimed (tested across the full contract registry: Section~\ref{sec:empirical-flip}).

\emph{Formal core (Lemma~A: disagreement-set upper bound).} For affine count aggregates with 1-Lipschitz outer map (accuracy; micro-/macro-F1 are ratio functionals and escape the premise, but micro-F1 is covered by a ratio-aggregate version with a measured $\gamma$-dependent constant, Lemma~A$'$, Appendix~B (supplementary)\footnote{The measured arms are resample quantities computed directly on the frozen grids, not lemma-derived bounds (E13 scope report); on the DocRE population the relevant denominators are thousands-scale (r61).}), $|\rho(M,c_1)-\rho(M,c_2)| \leq B\,|\mathcal{D}(g_1,g_2)|$ deterministically; $|\mathcal{D}(g_1,g_2)|/N$ for accuracy ($B{=}1/N$).  A model-dependent refinement: $|\Delta\rho| \leq 2\,\mathrm{arm}_{\mathcal{D}}\,|\mathcal{D}(g_1,g_2)|/N$, coupling the bound to prediction--disagreement overlap (the quantity Proposition~1 varies).  Coverage caveat: the flagship empirical object is a difference of sliced F1 readings, which neither Lemma~A (accuracy premise) nor Lemma~A$'$ (single micro-F1 reading) covers as stated; a per-slice extension of A$'$ (slice-level $\gamma$ and disagreement counts summed) is sketched but not claimed.  The bound is monotone in $|\mathcal{D}|$; the realized shift need not be (sign cancellation; Appendix~B (supplementary)).  \emph{Empirical surplus.} The family separation below is an ordering consistent with the envelope's slack structure (accuracy instance), not an ordering of the bound itself.

\emph{Assumptions.} Prediction-fixing levers ($s$, $g$, $\tau$) leave outputs unchanged; $f$, $\ell$ regenerate outputs and are treated separately.

\emph{Evidence and boundary.} Section~\ref{sec:empirical-flip} (Table~S.9 (supplementary): the family-level separation and two within-family slices are the supported content; open-ended generation is untested scope).

\subsection{Proposition 2: Rank-Flip Condition}
\label{sec:prop2}

\emph{Statement.}  For models $A$, $B$ and legal contracts $c_1, c_2$, a
\emph{rank flip} occurs when
$\operatorname{sign}(\rho(A,c_1)-\rho(B,c_1)) \neq
 \operatorname{sign}(\rho(A,c_2)-\rho(B,c_2))$.  \emph{Assumptions.} Arms and margins are computed on the same frozen predictions over legal contract pairs, under the document-paired resampling of \S\ref{sec:empirical-flip}.  A flip requires comparable margins (the necessary condition, Lemma~B below) and is empirically associated with long arms (low error-set agreement $A_{A,B}$): a far-apart pair does not reorder even with nearly disjoint errors.  Flips concentrate among statistically tied pairs: a dominant system shows level-arms without reversals (§\ref{sec:cross-task}), predicting where rankings move and where they cannot (human agreement similarly collapses on near-tie systems \citealp{hwang-etal-2026-deepqa}); all checks and Holm survivors: Section~\ref{sec:empirical-flip}.

\emph{Formal core (Lemma~B: no-flip margin guarantee).} Whenever $|\delta_{c_1}| > \mathrm{arm}(A)+\mathrm{arm}(B)$, the signs of $\delta_{c_1}$ and $\delta_{c_2}$ coincide: \emph{no flip is possible} (triangle inequality; proof: Appendix~B (supplementary)).  \emph{Empirical surplus.} Flips occur only where the margin is within the combined arms.  AER's protection is two-layered and the regions differ: tie declarations (CI-based, resample layer; empirically covering 14/14 registered flips) and the arm/margin audit (point layer, this Lemma's region); the former does not imply the latter.

\emph{Evidence.} Section~\ref{sec:empirical-flip}: thirteen flips among 57 checks, concentrated where the arm is long and signs contrastive; a leave-one-out check shows the condition carries flip-location information.

\subsection{Proposition 3: Variance--Level Ordering and a Seed Prescription}
\label{sec:prop3}

\emph{Statement.} The across-seed SD of a contrast is ordered by the level of randomness entering its effect path: V1 (scoring changes) $\leq$ V2 (decoding changes) $\leq$ V3 (training sources) $\leq$ V4 (double differences), strictness left to the data.
(A discretized generalizability-style decomposition~\citealp{cronbach1972dependability,brennan-2001-gtheory}; the four-benchmark test and the prescription it induces: Section~\ref{sec:empirical-llm}.)

\emph{Formal core (Theorem~C: variance ordering under nested randomization).} Nested effect paths with independent mean-zero increments give $\mathrm{Var}(\rho_k) = \sum_{i \leq k}\sum_{i \leq k}\mathrm{Var}(\varepsilon_i)$ by the law of total variance, hence $V_1 \leq \dots \leq V_4$ (proof: Appendix~B (supplementary)).  The theorem covers one contrast; ordering contrasts of different benchmarks additionally assumes comparable per-level increment magnitudes (tested alongside nesting and independence).
\emph{Empirical content.} The theorem fixes the assumptions the data must test: nesting and increment independence; the de-duplication cascade below is exactly the independence failure, and where the registered criterion fails.

\emph{Evidence.} Section~\ref{sec:empirical-llm} (Figure~\ref{fig:levelsd-main}, with load-bearing caveats: the points are not independent (the registered $\rho > 0.5$ fails the strictest de-duplication), and H3's benchmarks were substituted by amendment~6.

\emph{Corollary (seed prescription).} On the convergence curve of a HacRED decoding-side double difference (V4, 15 seeds; \S\ref{sec:empirical-llm}), the registered conservative track determines the sign at $n^{\ast}{=}8$, the exact track at $n^{\ast}{=}5$ (Figure~\ref{fig:seed-main}; a preliminary, contrast-specific count (per-contrast MDE analysis \citep{card-etal-2020-little}), conditional on equal-$\sigma$ imputation (10/15 seeds; pooled $\hat{\phi} = 0.34$, validated to $<3\%$) but insensitive to $\hat{\phi}\in\{0,\dots,0.80\}$ ($n^{\ast}{=}8$ throughout)).  The measured decomposition: between-seed standard deviations on the multi-seed arms ($.001$--$.047$) are negligible relative to the bootstrap half-widths for the fixed-contract arms (half-width is $1.2$--$1.8\times$ the between-seed SD) but dominant for the HacRED gold-bearing arm (between-seed SD is $5.1\times$ the half-width; the seed-inclusive interval no longer excludes zero at $n{=}5$), reported as sign-consistency across checks with disclosed seed sensitivity (r64/E-C3).  That arm is C3's own registered contrast (\S\ref{sec:empirical}): its Holm rejection is conditional on the frozen seed set, and the seed-inclusive interval excludes zero from $n{=}8$ (proportion $.991$) with the exact track reaching $n^{\ast}{=}5$ under equal-$\sigma$ imputation, the seed prescription is the reconciliation between the two statements.

\subsection{Proposition 4: Violations and Leverage (Assumed Independent. Observationally Decoupled)}
\label{sec:prop4}

\emph{Statement.}  Contract-violation rates (outputs failing structural
grounding) and leverage arms are independent dimensions of evaluation
validity (\S\ref{sec:empirical-llm}. LLM-gate boundary: \S\ref{sec:boundary}).

\emph{Assumptions.} Independence is assumed; the evidence is observational decoupling.

\emph{Evidence and boundary.} Section~\ref{sec:empirical-llm}: format changes move violations by $>$10 points and arms by ${\pm}0.003$; violations are graded; two independent-evidence settings come from gate-flagged cells, 16 of the 17 audited settings are gate-flagged (one passes by point estimate, none by CI, a non-vacuous bar: the CI pass ceiling is $3.0$--$4.1\%$ by cell size and the audited cells sit above it, r64/E-A12), whose violation rates limit generalization.  Observational decoupling under an assumption, not a theorem.

\looseness=-3
The three error sources (leverage bias, variance, violations) are validated separately below (anchor single-annotated; every segment-level claim carries that caveat, and the annotator facet is not modeled, the split-half floor is a lower bound on anchor noise). The decomposition is additive by construction with the cross-term vanishing empirically (r57); $z$-standardization removes per-contract location shifts by construction, so it captures variance, not bias: at segment level human--instrument disagreement accounts for effectively all error; at the system level, where rankings live, contract switches move a system's error by $0.013$--$0.043$ $z$ ($3$--$26\%$ of between-system variance), never dominating it; judge-side parse failures correlate with systematic bias ($+0.33$, fifteen cells): the selection bias the gate removes before marginalization.

\section{Empirical Validation}
\label{sec:empirical}

The account's assumptions and boundaries are empirical questions; this section tests them under the preregistered discipline.

\subsection{Setup}
\label{sec:setup}

\paragraph{Benchmarks and preregistration.}
\looseness=-3
All experiments follow a preregistration document frozen 2026-10-03
14:59 local (timestamped and sha256-pinned; the ledger timestamps every
later change; audit trail and verification records in the footnote~1
release), whose registration is provenance-explicit, e.g., the C3
seed grid is registered as three pre-freeze training runs plus ten
added after the freeze; the released bundle ships with sha256
manifests.  DocRED~\citep{yao-etal-2019-docred} contributes a 498-document dev subset with 189{,}556 frozen candidate pairs; its dev split carries 57.7\% cross-sentence gold links (58.2\% over the 498-doc retained subset), the slice we study. Two gold versions, gD and gR (revised projection of~\citealp{tan-etal-2022-revisiting}), instantiate the gold-revision lever (registry: Table~S.1 (supplementary); HacRED~\citep{cheng-etal-2021-hacred} (Chinese, 15 seeds) has one gold, its lever a decoding interaction; BC5CDR~\citep{li-etal-2016-bc5cdr} and RAMS~\citep{ebner-etal-2020-rams} are boundary benchmarks).

\paragraph{Predictors.}
\looseness=-3
Table~S.5 (supplementary) lists the seven families. X2 is an
instrument, a controllable prediction source, not a contributed
system.

\paragraph{Statistics.}
\looseness=-3
All intervals: document-paired bootstrap. The three comparisons were registered verbatim in the preregistration document frozen 2026-10-03 14:59 local (sha256-pinned; first ledger entry 15:01, and the ledger timestamps every later change); the primary confirmatory result is the Holm family over the two estimator-stable contrasts, C1--C2, both rejecting (adjusted $p$ $.0012$/$.0016$; r64/E-A12), with C3 retained in the registered family and demoted to preliminary below (its registered estimand and its Holm input disagree; the Holm(3) rejection is seed-conditional). (estimate [CI95]; adjusted $p$; power at the effect sizes shown / 80\%-power MDE, from the frozen power analysis.)  In plain terms on timing: the freeze locks the analysis protocol, not the data, the confirmatory tier C1's inputs are historical (existing, not rerun) and C3's first three training runs (2026-09-17/18/19) predate it, so the confirmatory tier is confirmatory in discipline rather than prospectively registered.  \textbf{C1}, registered as ``historical D-score (existing, not rerun)'': $D=(\mathrm{IIG}_{\mathrm{pair},D}-\mathrm{IIG}_{\mathrm{pair},R})-(\mathrm{IIG}_{\mathrm{gold},D}-\mathrm{IIG}_{\mathrm{gold},R})$, two scoring units (DocRED-trained vs.\ Re-DocRED-trained configuration; source subscripts D/R here denote training provenance, not the gold versions $g_D/g_R$) on the same frozen predictions of historical single-output models, five seeds averaged within resampling: $.036\ [.021,.050]$; $p{=}.0006$ (power $.79$ at effect $.020$. MDE $.020$).  The two scoring units are the positive-pair hit rate (A) and per-gold recall (R), $D=\Delta_A{-}\Delta_R$, each $\Delta$ the training-source contrast (five fixed historical models per source, estimand note in the bundle); the D-score shares the 498-doc frozen grid and the double-difference algebra with the registry's PE8$|$P3M gold flip ($.0363$, \S\ref{sec:empirical-flip}) but is a different contrast (scoring unit $\times$ training source, not model pair $\times$ gold version), the two near-coincident values are not independent evidence. \textbf{C2}, registered as ``P3M gold-swap (three existing $+$ two bridged seeds)'', confirmed on F1 (scoring-dependent: precision $.038\ [.017,.060]$, recall $-.007$; component-level interval not registered, point descriptive): $.039\ [.017,.060]$; $p{=}.0016$ (power $.44$ at effect $.020$. MDE $.031$). \textbf{C3}, registered as ``HacRED gold-bearing vs.\ fixed $\tau{=}0.5$, 15 seeds'': the decoding-side contrast, sign unstable across seeds (negative in 13/15), F1 component $+.015\ [.003,.027]$; $p{=}.011$ (power $.38$ at effect $.010$. MDE $.017$).  Two reconciliations the reader needs: the 13/15-negative profile is the $J$ statistic across the 15 decoding seeds, while the Holm-rejected quantity is the pooled F1 component over three seed points ($+.031/{+.031}/-.017$; the interval is carried by the two positive points), so the two signs describe different statistics (the three seed points are the three HacRED training runs, 2026-09-17/18/19, averaged within resampling under the same convention as C1; the registered 15 seeds are the decoding grid of the $J$ profile, r2); and C2--C3 are low-power contrasts, so their rejection is exact-test based and the magnitudes are read accordingly (C2's rejected magnitude $.039$ exceeds its MDE $.031$. C3's $.015$ falls below its $.017$).  Per-seed sensitivity: C2's F1 component is positive in all three seed points ($.040/.034/.041$), sign-stable. C3's F1 component flips sign ($.031/.031/-.017$), C3 is therefore reported as \emph{preliminary at its own MDE} wherever cited.  Family sensitivity: Holm over the remaining two (C3 removed) still rejects both (adjusted $p$ $.0012/.0016$; r64/E-A12).  C3's specification, put concretely: the registered clause names the $J$ estimand over a 15-seed decoding grid, while the freeze's primary-metric section designates micro-average F1, the Holm input is that F1 component pooled over the three training-run seed points, so C3's rejection is \emph{seed-conditional} wherever cited, and its seed-inclusive status is carried by the prescription curve (\S\ref{sec:prop3}: exclusion probability $.991$ at $n{=}8$).

\subsection{Leverage and Rank Flips}
\label{sec:empirical-flip}

\looseness=-3
Four questions structure this section (one per proposition; the flip registry is \emph{exploratory} throughout; the confirmatory layer is C1--C3, above); four checks rule out slice selection, grouping, format, and population; three probes carry the axes beyond DocRE (arms: the twelve-setting pool; census: the seventeen LLM settings).

\paragraph{Generator-dependent arms (Proposition~1).}
\looseness=-3
Across twelve generator settings,\footnote{Eleven measurable-density settings plus the registered-only Qwen-json arm; 57 checks/21 pairs.}, within-model format changes move arms by at most ${\pm}0.003$, family-level separation tracks prediction--disagreement overlap rather than capability (7B$\to$14B lowers both), and density tracks the arm as a tendency, not a law, with two inversions precluding monotonicity (density--arm Spearman $+0.64$; Table~S.9 (supplementary)).

A training-investment confound remains, bounded but not excluded by three checks (supplement): a matched-F1 pair (KD-DocRE and DREEAM at the same dev F1, arms differing by 0.012), a Mann--Whitney separation of weak- from strong-configuration arms ($p{=}0.029$), and second-seed replication of the arms, each consistent with architecture-driven arms, none excluding the shared quick-retrain protocol.

\begin{figure}[htbp]
\centering
\includegraphics[width=0.98\linewidth]{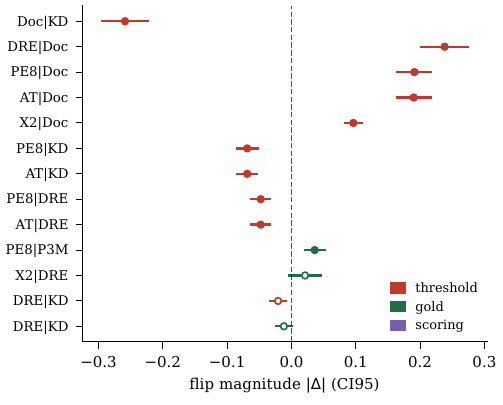}
\caption{The thirteen registry flips (Proposition~2): signed flip delta with
CI95, by axis; filled markers survive Holm (10 of the 11 CI-excluding).
Flips concentrate on the threshold axis and among near-tied pairs.  Pair
abbreviations: PE8 = ATLOP-PE8, AT = ATLOP-full, DRE = DREEAM,
Doc = DocuNet, KD = KD-DocRE.}
\label{fig:flipsforest}
\end{figure}

\paragraph{Rank flips (Proposition~2).}
\looseness=-3
 Among 57 contract-pair checks over twenty-one encoder-family pairs, thirteen flips occur, eleven of them CI-excluding (Figure~\ref{fig:flipsforest}; full registry: Table~S.10 (supplementary)).  These are checks, not thirteen independent events: systems and the fixed-vs-decoded contrast are shared across rows.  \begin{theorem}[Local-Action Algebra]
\label{thm:local-action}
Let $c_1, c_2$ be two evaluation contracts that act locally on a frozen
prediction set: there exists an enumerable set $S(c_1, c_2)$ such that
all count changes between $\rho(M, c_1)$ and $\rho(M, c_2)$ are
confined to $S$.  On the gold axis, $S$ is the gold-difference region
$S_a \cup S_r$ where $S_a = g_2 \setminus g_1$ and
$S_r = g_1 \setminus g_2$; on the threshold axis, $S$ is the
prediction-set symmetric difference $P_1 \triangle P_2$.  In both
cases, the reading difference $\rho(M, c_1) - \rho(M, c_2)$ is an
exact function of the status tally of $S$: the partition of $S$ into
(gold-positive $\times$ predicted-under-$c_1$-only /
predicted-under-$c_2$-only / both / neither) cells.
\end{theorem}
\begin{proof}
We derive the gold axis; the threshold axis follows by the same partition
argument with gold fixed and prediction membership varying.

Let $h_a = |P \cap S_a|$ and $h_r = |P \cap S_r|$.  By locality,
predictions outside $S_a \cup S_r$ retain their TP/FP/FN status under
both contracts.  Consider $p \in S_a$.  Under $c_1$ (gold $g_1$),
$p \notin g_1$, so $p$ is a false positive if $p \in P$ and a true
negative if $p \notin P$.  Under $c_2$ (gold $g_2$), $p \in g_2$, so
$p$ is a true positive if $p \in P$ and a false negative if
$p \notin P$.  Each element of $P \cap S_a$ therefore moves from FP
to TP. Each element of $S_a \setminus P$ moves from TN to FN.  Summing:
\begin{equation}
\label{eq:rho}
\mathrm{TP}_2 = \mathrm{TP}_1 + h_a, \quad
\mathrm{FP}_2 = \mathrm{FP}_1 - h_a, \quad
\mathrm{FN}_2 = \mathrm{FN}_1 + (|S_a| - h_a).
\end{equation}
The case $p \in S_r$ is symmetric: elements of $P \cap S_r$ move from
TP to FP, elements of $S_r \setminus P$ move from FN to TN.  Combining both sides (\eqref{eq:rho} and~\eqref{eq:localaction}):
\begin{align}
\label{eq:localaction}
\mathrm{TP}_2 &= \mathrm{TP}_1 + h_a - h_r, \\
\mathrm{FP}_2 &= \mathrm{FP}_1 - h_a + h_r, \\
\mathrm{FN}_2 &= \mathrm{FN}_1 + (|S_a| - h_a) - (|S_r| - h_r).
\end{align}
Since $\rho$ is a function of the count vector
$(\mathrm{TP}, \mathrm{FP}, \mathrm{FN})$, the reading difference is
determined by $(h_a, h_r, |S_a|, |S_r|)$; the status tally of $S$.
\end{proof}

\begin{remark}[Empirical verification]
The identities above were verified by exact per-document recount on all
five stand-in systems: algebraic recounts match direct recounts
bit-for-bit (r64/E-A16 for the gold axis, r64/E-A19 for the threshold
axis).
\end{remark}

Mechanistically, the two axes carry different engines.  On the gold axis, the Re-DocRED authors themselves identify the engine, models predicting true relations missing from gD are charged false positives, so high-recall designs pay a systematic tax \citep[cf.~][]{tan-etal-2022-revisiting}, and this audit adds what their retraining-based comparison cannot show: with predictions frozen, the remittance is uneven and gap-specific.  The 10{,}887 relations added on the audit subset are themselves 70.8\% cross-sentence, so remitting the tax reshapes intra/inter weights per model rather than uniformly, and a flip occurs where the remittance difference exceeds the pre-switch near-tie (the rank-flip condition of \S\ref{sec:formal}).  The decomposition is also predictive: the switched counts are an exact algebraic function of the hit ledger alone (identity verified on all five stand-ins), added-region hits span 1{,}344--5{,}854 across the audited systems, and the headline reversal would require 237 additional added-region hits for DREEAM, a deterministic analogue of the confidence interval (r64/E-A16).  A per-class recall-profile extrapolation over the added-region composition overpredicts the remittance by a difficulty premium of 68--74\% on four of the five stand-ins (consistent across models; the memory-augmented TTM-RE, the C3 comparator, shows only 39\%), direct quantitative evidence that the under-annotated stratum is a hard, non-random subset (r64/E-A17).  Leave-one-system-out calibration makes the prior declarable rather than in-sample: within the standard-architecture class the premium transfers out-of-sample (error 0.2--14.6\%, median 8.7\%), while the mixed pool fails (median 32.4\%), the memory-augmented model is a labelled failure mode, not an unexplained one (r64/E-A18).  On the threshold axis, each published number was optimized at its own operating point, and re-scoring at one common threshold re-allocates credit asymmetrically across score-distribution shapes.  The threshold axis admits the same local-action algebra: a threshold switch moves counts only within the symmetric difference of the two prediction sets, and on the raw-logit space that difference spans two orders of magnitude across systems and seeds (13.8k to 5.9M predictions, while its gold-positive core stays near 3.3--5.2k), re-scoring ATLOP-family logits at a fixed sigmoid 0.5 costs 47--61 F1 points against the native decode, which is why adaptive thresholding was invented (r64/E-A19).  One sort of the logits gives the whole fixed-threshold family's audit exactly, and the curves split the systems into two logit semantics: probability-calibrated (DocuNet \citep{chen-etal-2021-docunet} peaks at $t{=}{-}.25$ with $F_1$ $0.591$ at $t{=}0$) and ranking-calibrated (the ATLOP family collapses to $F_1$ $0.004$--$.018$ at $t{=}0$ and rises monotonically to $.43$--$.44$ at the grid edge), no single scalar threshold is a common operating point, and pair-level flips inside the family have computable locations with bootstrap uncertainty bands (r64/E-A20).  Whether the gold-side remittance is artifact correction or construct drift is deliberately not adjudicated (\S\ref{sec:aer}); the mechanism here is what makes either reading possible.  The mass concentrates on the threshold axis: all ten threshold-axis flips involve the fixed-$\tau{=}0.5$ vs.\ decoded-threshold contrast, a policy choice some systems optimize internally; how widely the two policies occur in published practice was sampled for this revision: in a twelve-paper sample of published DocRE model evaluations (the audited systems' sources plus canonical models, 2021--2025; eleven full texts retrieved, one PDF not retrievable after two attempts), seven of the ten classifiable papers report under adaptive or learned thresholds, one (DocuNet) under a fixed zero-margin (sigmoid $0.5$), and two are silent in text (r64/E-A13); headline numbers across papers do not share one threshold policy, while the mass remains one contrast instantiated ten times rather than ten independent contract ambiguities.  The remaining three sit on the gold axis: the ATLOP-PE8 vs.\ P3M reversal previewed in Section~\ref{sec:intro} (CI-excluding, Holm-surviving), plus two CI-including flips; the scoring axis contributes none.

The pattern matches the leverage condition (longest threshold arms: KD-DocRE, DREEAM, ATLOP variants; calibration \emph{raises} DocuNet's gap while \emph{collapsing} KD-DocRE's) and modestly exceeds redescription: a single-threshold classifier on arm length reaches leave-one-out accuracy $0.860$ vs.\ majority $0.772$ ($p<0.002$).  Under Holm 34 of 57 survive, as do 10 of the 11 CI-excluding flips. Kendall $\tau \in [-0.333, +1.000]$ across the registry's 15 contract dimensions (supplement), contract identity, not only effect magnitude, is unstable across dimensions.

\paragraph{Auditing published orderings.}
\looseness=-3
The same machinery re-scores three \emph{published} claims under both golds, through retrained stand-ins for the published systems (published-configuration checkpoints land $-3.6$ to $-10$ points below published numbers through their own entrypoints; Appendix~E (supplementary); Table~\ref{tab:claims}\footnote{Bundle artifact r56, gold-axis recomputation v2.}): one of three reverses under the revised gold, and the audit separates contract from population effects.  Every verdict is stable across all available seeds\footnote{Per-seed recomputation, bundle artifact r56/PS.} (claim rows CL1--CL3 here, not to be confused with the confirmatory comparisons): CL1 reverses in 3/3 DREEAM seeds (gR $\Delta$ $-.018$ to $-.025$, each CI-excluding), CL2 holds in 2/2 aligned seed pairs (gD $+.146/{+.096}$, gR $+.050/{+.039}$, all CI-excluding), CL3 reads opposite-side in 2/2 (gD $-.065/{-.032}$, gR $+.290/{+.293}$), a population verdict, not a gold one: the published ordering already fails to reproduce on the audit population under $g_D$ ($-.048$), so CL3 is reported as population-driven and kept out of the gold-flip count.  The CL1 divergence is convention-robust: under a second independent ATLOP retraining convention, all three seeds reproduce the gD-above/gR-below split, each side CI-excluding (r64/E-A10).  Stand-in fidelity: the retrained checkpoints reach full-dev best F1 $.592/.580$ (DocuNet, 2 seeds), $.610/.577$ (KD-DocRE), $.622$ (DocuNet-large), with the audit itself running on the 498-doc subset (F1 $.577$--$.613$, Table~S.5 (supplementary)); the published-artifact failure audit that motivates retraining is Appendix~E (supplementary).  Positioning against the retrained regime: DREEAM's own comparison table reports the post-2022 systems retrained on the revised corpus (Re-DocRED test-set scores under RoBERTa-large encoders: ATLOP $77.56$, DocuNet $77.87$, KD-DocRE $78.28$, DREEAM $77.94$; \citealp{ma-etal-2023-dreeam}): under retraining, DREEAM$>$ATLOP ($+0.38$) and KD-DocRE$>$DocuNet ($+0.41$) are preserved, while the frozen audit reverses CL1 ($-.0215$).  The two regimes are not commensurable as published (retrained test-set scores vs.\ the 498-doc frozen revised-dev subset), but the divergence is itself informative: a retrained comparison rebuilds the pipeline under the new gold and cannot show which published near-ties are gold-fragile at prediction level, which is exactly what the frozen audit measures.  The designated follow-up has been executed (r64/E-A24): ATLOP and DREEAM stand-ins were retrained on the revised training corpus under the same PE8-aligned exclusion (3 seeds each, 30 epochs, evaluated on the same 498-doc revised dev; checkpoints selected by best dev F1 over epochs, matching the training protocol's model-selection rule).  Under retraining, DREEAM$>$ATLOP is preserved: the seed-aligned margins are positive in two of three pairs ($+0.66/{-}0.10/{+}0.87$ points), and the pooled document-paired bootstrap margin over the 498 revised-dev documents is $+0.48$ points with 95\% CI $[+0.07, +0.88]$, excluding zero (10{,}000 resamples; both arms are BERT-base under identical training conditions, so comparisons to the literature values are directional only; r64/E-A24), so the retraining confound does not explain the frozen reversal, and the regime divergence stands as the paragraph's opening described.  The published margin ($+0.33$pt), the retrained margins ($+0.38$ and $+0.48$), and the frozen margin ($-0.0215$) all sit inside the near-tie band where Lemma~B locates flip risk.  Verdict transfer: CL1's published margin is itself small, DREEAM (BERT teacher) dev F1 $61.42$ vs.\ ATLOP (BERT-base) $61.09$, $+0.33$pt \citep{ma-etal-2023-dreeam,zhou-etal-2021-atlop}, and the stand-in gD margin ($+.61$pt) matches it in sign and order of magnitude; by Lemma~B, flips occur where margins fall inside the combined arms, so a verdict established at a small margin is not sensitive to stand-in fidelity of this size, and CL2/CL3's larger margins ($+.121$, gR-side $+.292$) sit far outside any plausible fidelity gap.  The official artifacts cannot serve this re-scoring themselves: rerun through their own entrypoints, both published-config checkpoints degrade comparably (teacher $-10.1$, BERT-base $-8.4$; Appendix~E (supplementary)) and their rerun margin \emph{inverts} the published ordering (ATLOP 52.7 above DREEAM teacher 51.3), while the stand-ins preserve it, degraded artifacts reorder near-ties; matched stand-ins do not.  The stated assumption: verdicts transfer insofar as stand-in margins preserve the published near-tie structure, which these numbers support but do not prove.  A perturbation audit makes the tolerance empirical: TP-removal degradation of either stand-in up to 2\%, beyond the observed stand-in--published gaps, flips no verdict sign and loses no CI exclusion, and the headline CL1 reversal breaks only at 4--8\% comparator-side degradation, several times any observed gap (r64/E-A14).  The battery is extended symmetrically (r64/E-A22, released with the bundle): FP-injection up to 8\% of per-document TP, on either side of the CL2 pair, moves the gD margin by at most $0.0082$ and never unseats the verdict (the CI excludes zero in all 16 injection settings; side-neutral field semantics in the released artifact), and cross-document redistribution of errors with totals preserved leaves the aggregate margins identical in 10/10 repetitions: aggregate micro-F1 is redistribution-invariant by construction, which is precisely why the sliced readings of \S\ref{sec:empirical-flip}, not aggregates, carry the distributional evidence.  On the headline CL1, DREEAM-side FP-injection deepens rather than erases the gR reversal ($-.0215\to-.0267$ at 8\%), and moves the gD margin across zero only at 8\% ($+.0061\to-.0022$), beyond any observed stand-in--published gap; with the TP-removal results, the claims-audit verdicts now sit inside a two-direction perturbation envelope (E-A14 + E-A22).

\begin{table}[t]
\centering
\caption{Three published orderings re-scored under both golds, with frozen predictions from retrained stand-ins.}
\label{tab:claims}
\setlength{\tabcolsep}{2pt}
{\scriptsize
\begin{tabular}{@{}llll@{}}
\toprule
Claim & gD & gR & Verdict \\
\midrule
DREEAM$>$ATLOP & $+.0061$ & $-.0215$ & reversed (gold) \\
KD-DocRE$>$DocuNet & $+.121$ & $+.044$ & holds, arm $\div 3$ (gold) \\
TTM-RE$>$KD-DocRE & $-.048$ & $+.292$ & opposite (population) \\
\bottomrule
\end{tabular}
\par\vspace{3pt}\footnotesize
\emph{Notes:} Paired reading differences are instantiated at the F1-based scorer under each stand-in's native decoding convention (the leverage-arm convention of Table~S.9 (supplementary)); the registry's fixed-$\tau{=}0.5$ contrast, \S\ref{sec:empirical-flip}, is a stressor instantiation, not this table's operating point. Per-seed CI95s in the text (r56/PS); reversed = gD tie, gR CI-excluding. Claims from \citealp{ma-etal-2023-dreeam,tan-etal-2022-document,gao-etal-2024-ttmre}. Driver: \emph{gold} = the verdict change is attributable to the gold switch (population shared); \emph{population} = the published ordering fails to reproduce on the audit population already under gD ($-.048$), so the gR-side flip is not a gold effect. The reversed verdict is marked in words in the Verdict column.
}

\end{table}

\subsection{Variance Levels, Violations, and Further Axes}
\label{sec:empirical-llm}

\begin{figure}[htbp]\centering
\includegraphics[width=.8\linewidth]{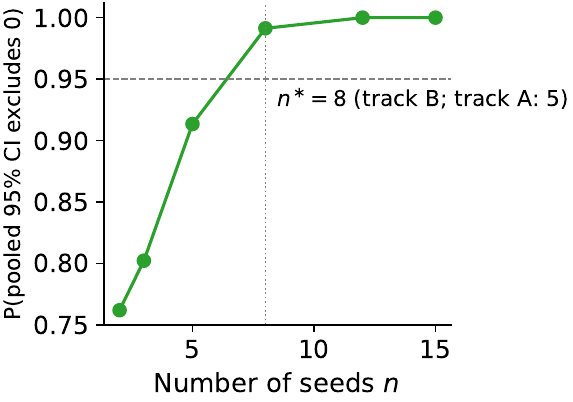}
\caption{Seed prescription for the HacRED decoding-side double-difference contrast: fraction of $n$-seed subsets whose pooled 95\% CI excludes zero. The preregistered 95\% criterion is met at $n^{\ast} = 8$ under the conservative track (equal-$\sigma$ imputation for 10/15 seeds); the exact document-bootstrap track gives $n^{\ast} = 5$.}
\label{fig:seed-main}
\end{figure}

\begin{figure}[htbp]\centering
\includegraphics[width=.85\linewidth]{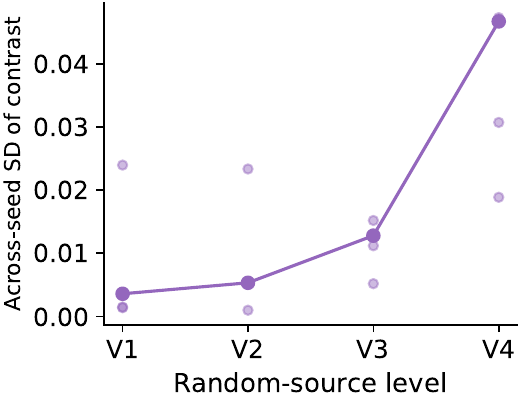}
\caption{Across-seed SD of 15 contrast points from four benchmarks, by random-source level (V1--V4). Lines connect the upper-median point per level; higher levels introduce more randomness into the reading; the ordering Proposition~3 formalizes. The pooled association is $\rho = 0.656$, $p = 0.0045$, reported as secondary because the points are not independent; under the strictest de-duplication $\rho = 0.359$ ($p = 0.27$, $n = 6$).}
\label{fig:levelsd-main}
\end{figure}

\paragraph{Variance--level ordering (Proposition~3).}
\looseness=-3
The pooled 15-point level-vs-SD Spearman is $r_s = 0.656$ (one-sided
permutation $p = 0.0045$), with monotone level medians across V1--V4
(Figure~\ref{fig:levelsd-main}).  The points
are not independent, however, and nine are calibration-derived, so the
pooled test is secondary.  De-duplication gives $r_s = 0.588$
($p = 0.017$), $r_s = 0.493$ ($p = 0.044$), and $r_s = 0.359$
($p = 0.27$, six non-calibration points): the registered $r_s > 0.5$
pass does not survive the strictest.  Per benchmark the criterion is
unmeetable by construction (RAMS $\rho = 0.5$; BC5CDR minimum
attainable $p = 0.17$ at $n = 3$): a failed registered
sub-hypothesis.
Within HacRED the 15-seed double difference shows the prescribed
convergence (pooled-CI exclusion reaches $0.991$ at $n = 8$, Figure~\ref{fig:seed-main}; exact
track $n^{\ast} = 5$), while the X2 source contrast stays
sign-undetermined at $n = 10$.  The subset machinery replicates on the X2 family's second
contrast (criterion at $n{=}5$ for $\phi{=}1.0$; undetermined for
$\phi{=}0.25$ at all $n{\leq}10$; r40).

\paragraph{Per-seed sign stability (preregistered H1).}
\looseness=-3
The preregistration predicted sign consistency below $0.8$ at 15 seeds; the observed $0.867$ (13/15 negative) contradicts the clause in the point estimate, though not significantly ($P(X{\ge}13 \mid p{=}0.8) = 0.40$; diagnostics: supplement).

\begin{figure}[htbp]\centering
\includegraphics[width=.85\linewidth]{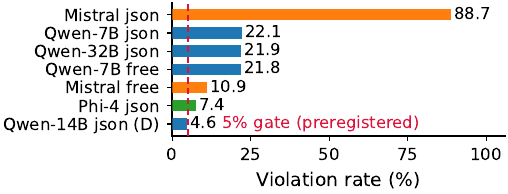}
\caption{Violation surface (Proposition~4): contract-violation rates (\%) over seven instruction-tuned settings: six Chinese HacRED cells plus the English-DocRED cell of Qwen-14B-json (D), the one setting passing the gate (its Chinese value: 28.9); the dashed line is the preregistered 5\% gate (full seventeen-setting census: supplementary Table~S.2).}
\label{fig:vio-main}
\end{figure}

\paragraph{Violation surfaces under LLMs (Proposition~4).}
\looseness=-3
Over the instruction-tuned grid (two models $\times$ two formats, HacRED, NF4 quantization, greedy; extension adds 32B), three patterns matter.  (i) \emph{The failure surface is a (model $\times$ instruction language $\times$ format) object}: same-dataset controls give opposite answers for the two models: Qwen-7B barely moves between prompt languages, Mistral-7B swings $8.1\times$ between formats (Figure~\ref{fig:vio-main}; Table~S.2 (supplementary)); the inventory channel is dataset-borne (1.7\% English-DocRED vs.\ 13.8\% English-instruction HacRED, same model; channel shares per the single-coder taxonomy of \S\ref{sec:audit}).  A doc-clustered decomposition attributes incidence to relation frequency (odds ratio [OR] $0.43$) and model identity, not format (OR $1.02$).

\looseness=-3
Mistral-7B collapses in exactly one cell (Chinese json), re-read at 55.0\% under a stop-sequence constraint: the violation reading itself is decoding-contract-relative.  The two failure channels separate: collapse cells are parse-failure-dominated, while the six HacRED-side Qwen cells carry 55.9--78.1\% of violation mass in the label-inventory channel: survivor-set arms therefore compare parseable populations.  Selection on parseability is itself a contract effect: survivor-set arms are conditional on survival, fractions per cell: supplement.
(ii) \emph{Violations are a (model $\times$ contract) interaction}: format relaxation rescues the collapsing cell, language switching roughly halves it, Qwen-7B unchanged (Table~S.2 (supplementary)).  (iii) \emph{Violations and leverage move independently}: the lowest-violation setting carries a near-zero arm; 7B$\to$14B lowers violations (11.4\%$\to$4.6\%) and the arm ($+0.0174\to+0.0063$); the Chinese counterpart rises ($\Delta{+}6.7$pt, CI $[+3.8,+9.7]$), carried by the out-of-inventory name channel (14.4\%$\to$22.6\%, offset by a decline in action-form slips, 23\%$\to$4\%; decomposition in the appendix): the larger renames the code, the smaller slips morphologically.

\begin{figure}[htbp]
\centering
\includegraphics[width=\linewidth]{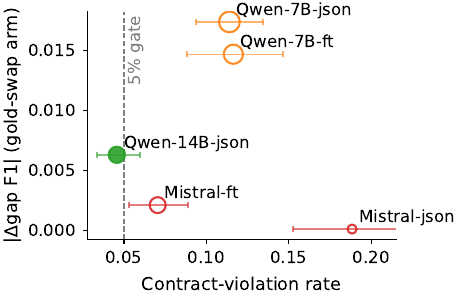}
\caption{Contract-violation rate vs.\ gold-swap arm for the five
instruction-tuned settings with a gold pair (English DocRED prompts;
bubble size $\propto$ mapped triples); open markers: flagged unverifiable
by the preregistered 5\% gate (four of five; the passing cell reads 4.57\%).  The four Chinese HacRED
cells have no gold pair (rates: Figure~\ref{fig:vio-main}; full census: Table~S.2 (supplementary)).}
\label{fig:quad}
\end{figure}

\paragraph{The 5\% gate, applied uniformly.}
\looseness=-3
The gate marks a setting above 5\% unverifiable: the conventional Type-I rate (verdicts at 3\%/10\% alongside, Table~S.2 (supplementary)); by the per-pair counting bound (micro-accuracy), survivor-set selection bias is at most $v\,|a_{\mathrm{surv}}-a_{\mathrm{viol}}| \leq v$. Of seventeen settings one passes by point estimate; none passes a CI gate (Table~S.2 (supplementary)); the extreme cell maps 192 of 1{,}701 triples, unreliable under every threshold column.  A post-review stop-sequence diagnostic (amendment~82) localizes that extreme cell: constraining Mistral-7B in the collapsing cell to halt at the few-shot separator (zero truncations, 1{,}500/1{,}500) re-reads it at $54.97\%$: roughly a third of the collapse is unstopped-truncation artifact, and the residual is parse-failure-dominated (1{,}245 of 2{,}507 events); the character-vs-word violation-rate gap reads $+0.0003$ with L2 difference $0.000$.  Gated settings therefore supply descriptive evidence only (open markers in Figure~\ref{fig:quad}), not confirmatory; smoke runs do not extrapolate.

\paragraph{Scale and family extension (preregistered bands).}
\looseness=-3
The preregistered extension (amendment~85) fixed two bands (15--35\% = Qwen-profile replication, below 10\% = capability-sensitive). Both occur across families rather than scales (32B Qwen scale-robust, same-scale Phi-4 at $7.4\%$, family $\times$ dataset-bound; Table~S.2 (supplementary)); quantization replicates, with every quantized cell above the gate.

\paragraph{The grouping rule itself (registered H-G1).}
\looseness=-3
The intra/inter split is itself a contract coordinate: recomputing every run under \texttt{any\_shared\_sentence} (ASS) vs.\ frozen \texttt{earliest\_mention} (ESS) moves the readings of all eight generators with stored pair-level logits (every CI excluding zero, arms exceeding the gold-swap arm for all three public retrains; H-G1 PASS; five generators in Table~S.11 (supplementary), three post-review levers in Table~S.12 (supplementary)). The direction is uniform (ASS shrinks the gap in every run), so no pair flips: a long arm with non-contrastive signs moves levels, not rankings (X2/P3M lack pair-level scores).

\paragraph{Replication on an unseen population (gold axis).}
\looseness=-3
The gold-swap protocol replicates on a second same-source population: the 500 Re-DocRED test-revised documents (schema-identical to the dev release). Five families have retained checkpoints: the three dev-significant levers replicate at the point-estimate level (DocuNet-L $0.025\to0.041$; TTM-RE same-signed in both seeds); near-zero arms stay near zero; a published-configuration RoBERTa-large ATLOP enters as a near-zero lever ($0.021$, CI $[-0.048,+0.005]$): the DREEAM profile at large scale, across five provenance tiers.

\paragraph{Slices and robustness.}
\looseness=-3
Robustness slices (relation frequency, document length, sentence distance) show the gap positive in all 91 slice tables across seven arms; population slices leave the sign fixed while contract axes move it, and UD label-revision slices localize movement to the re-annotated class (Table~S.14 (supplementary): slice choice does not carry the movement).

\subsection{Cross-Task Replication}
\label{sec:cross-task}

\looseness=-3
Six further task families replicate the audit under the same frozen-prediction discipline: retrieval on TREC DL 2019 \citep{craswell-etal-2020-trecdl} (r49), MT on WMT19 \citep{barrault-etal-2019-wmt19} (r50), QA on SQuAD v2 \citep{rajpurkar-etal-2018-squad2} (r51), code (r52), and the released-gold audits (r59/N1--N2); observational breadth for Props.~1--2, outside the preregistered ledger (readings and flip counts per family: Table~\ref{tab:multitask}).  The UD row is two objects: released-parser re-scoring under the version gauge (frozen predictions) and within-regime training-contract probe flips (not frozen across regimes; Table~S.14 (supplementary)).

\begin{table*}[t]
\centering
\caption{Cross-task replication summary: frozen predictions, one contract switch at a time.}
\label{tab:multitask}
\setlength{\tabcolsep}{6pt}
{\small
\begin{tabular}{@{}lll@{}}
\toprule
Family / axis & Reading & Arm or flips \\
\midrule
Retrieval: gold rubric & nDCG@10 & $+0.137$--$0.158$ (5 runs) \\
  (graded $\to$ binarized) & MAP@100, bpref & $0$ (by construction) \\
MT de-en: scorer & BLEU$\to$chrF & 3/120 pairs flip \\
  (16 frozen systems) & BLEU$\to$BERTScore & 10/120 flip, $\tau{=}0.83$ \\
MT: format & BLEU 13a$\to$none & $\leq 5.3$ BLEU \\
QA: gold population & F1 & $0.189 \to 0.378$ \\
  (v2.0 $\to$ v1.1-style) & EM & $0.096 \to 0.192$ \\
QA: scorer & EM$\leftrightarrow$F1 & $\approx 0.10$; 1 pair flips \\
QA: threshold & EM vs.\ F1$\geq$0.5 & $\approx 0.12$ \\
Code: gold tests & pass@1 & HE $0.835 \to 0.768$ \\
  (base $\to$ EvalPlus) & pass@1 & MBPP $0.794 \to 0.656$ \\
  (3 coder models) & ranking & no flip. Qwen leads both golds \\
Code: format & fenced vs.\ raw & up to $0.84$ of pass@1 \\
Code: decoding & greedy vs.\ sampled & $\leq 0.018$ (n.s.) \\
\bottomrule
\end{tabular}
\par\vspace{3pt}\footnotesize
\emph{Notes:} Retrieval: five frozen runs, TREC DL 2019 (43 topics); zero flips in 66 gold$\times$scorer pairs. MT: WMT19 (16/15 systems); newstest 2021--23, eleven into-English grids of 8--21 systems, consecutive-scorer flip rates 5.0\% (tokenizer), 7.7\% (chrF), 14.1\% (chrF$\to$BERTScore), full matrix 5.0--14.8\%; 22 released En$\to$De submissions under five human-written alternative reference sets (AR, AR.p, WMT.p, HQ(R), HQ(P), the alternative reference sets of \citealp{freitag-etal-2020-bleu}, labels as in our bundle artifacts; 21 with usable output) flip 12--24/210 pairs per scorer (r59/N3, E1). QA (question answering): three extractive models, SQuAD v2 dev. Code: three coder models, HumanEval/MBPP \citep{chen-etal-2021-codex,austin-etal-2021-mbpp} vs.\ EvalPlus \citep{liu-etal-2023-evalplus}, frozen samples $n{=}5$--$10$. RE: 45 released TACRED predictions, TACREV/Re-TACRED gold gauges (subset-matched re-scoring $-13.7$ to $-21.6$, N1b). MCQA: five frozen models from three families, original vs.\ Redux-corrected labels (r59/N2). UD: two released UDPipe parsers \citep{straka-strakova-2017-udpipe}, $+0.045$/$+0.082$ LAS under the v1$\to$v2 gauge switch over Universal Dependencies \citep{nivre-etal-2016-ud,nivre-etal-2020-ud} (r59/E3); AER-vs-single TOST and flip coverage (r59/E2), flip counts there are within-regime training-contract arm families, not frozen across regimes (Table~S.14 (supplementary) caption); gauge-breaking scorer $-0.003$ under pure rename (r59/E5).
}

\end{table*}

\bigskip

\looseness=-3
Three readings carry the section.  First, \textbf{gold-coordinate switches move readings wherever they are legal} (Table~\ref{tab:multitask}. MAP/bpref exactly invariant under rubric binarization, r58), and the two TACRED revisions move the leaderboard leaders themselves \citep{alt-etal-2020-re-tacred,stoica-etal-2021-retacred}: over 45 released systems, 88/990 pairs flip between TACRED and its same-guideline re-annotation TACREV, CI-tie coverage 88/88 (r64).  (TACREV's paper reports 49 baselines; its released prediction set resolves to 45 usable systems, on which this audit runs, r59/N1.)  Second, \textbf{the MT scorer and tokenizer axes persist} \citep{post-2018-sacrebleu} across newstest 2021--23 and the released alternative reference sets \citep{freitag-etal-2020-bleu}; our audit replicates it (r59/E1).  Third, \textbf{flips concentrate where margins are near-tied} (Table~\ref{tab:multitask}; 40 of the 88 TACRED flips sit inside the $\pm0.5$pp slice) and far-apart pairs mostly shift levels; the output-format contract on code is larger still (up to $0.84$ of pass@1).  AER transfers to these grids as declared: retrieval (four contracts) and code show zero contract-induced flips, the QA EM$\leftrightarrow$F1 flip is ensemble-covered (r59/N4).

\section{From Falsification to Construction: The Audited Evaluation Reading}
\label{sec:aer}

\looseness=-3
The \emph{audited evaluation reading} (AER) is the constructive response to the falsifications above, with four components:
\textbf{(i) marginalization}: report the registered contract-set average $\bar\rho(M) = \mathbb{E}_{c\in\mathcal{C}}\, z_c(\rho(M,c))$, $z_c$ the within-contract standardization over systems;
\textbf{(ii) tolerance}: a pair is \emph{tied} when a segment-bootstrap CI of the paired difference contains zero; orderings are claimed only for CI-excluding pairs (touching counts as tied);
\textbf{(iii) prescription}: the seed and contract sample sizes
needed for a target precision (Section~\ref{sec:prop3});
\textbf{(iv) gate}: readings failing the validity gate are excluded,
using the argumentation reading of that word, an appraisal step in a
validity argument rather than a binary truth claim
\citep{kane2013validity} (Section~\ref{sec:empirical-llm}; tooling in
Section~\ref{sec:audit}).
The leverage law applies to AER itself: leave-one-member-out moves its reading in $[0.667, 0.700]$ (de-en) and $[0.752, 0.771]$ (zh-en; there the chrF$+$BERTScore pair, $0.771$, dominates the registered ensemble, $0.752$; r55): member selection is itself a contract choice. AER claims are relative to the declared, frozen member set.

\looseness=-3
On the WMT19 de-en and zh-en grids (r60 canonical re-freeze; contract set \{BLEU-13a \citep{papineni-etal-2002-bleu,post-2018-sacrebleu}, chrF \citep{popovic-2015-chrf}, BERTScore \citep{zhang-etal-2020-bertscore}\}; $R{=}2000$ segment resamples, ties re-declared at $R{=}2000$; anchor = official raw human scores).\footnote{Estimator calibers: \emph{point} = original-sample $\tau_b$ of the $z$-ensembled reading. CIs = percentile intervals over resamples; $\Delta$ = point $\tau_b$(AER) $-$ point $\tau_b$(best single), the comparator held fixed across resamples (no in-resample re-selection; the $\Delta$ interval conditions on that choice).}  The ensembled reading agrees with the anchor at point $\tau_b$ $0.700$ on de-en (single contracts $0.683/0.667/0.650$) and $0.752$ on zh-en (singles $0.733/0.752/0.771$), while carrying tie declarations.  Equivalence to the best single contract holds at $\epsilon{=}0.10$ on both grids (two one-sided tests, TOST, \citealp{schuirmann1987approximate}. Table~\ref{tab:aerties} carries $\Delta$, all intervals, and the full grid).  Margin provenance and stability: neither margin was in the 2026-10-03 freeze, $\epsilon{=}0.05$ enters at the revision round, and the $\{0.05,0.10\}$ grid with a CI-method scan is exploratory; across that scan, $\epsilon{=}0.10$ passes on both grids under both percentile and basic intervals at every $R\in\{200,1000,2000\}$, whereas $\epsilon{=}0.05$ is convention-dependent (the $R{=}200$ freeze recorded de-en as not passing, lower limit exactly $-.050$, boundary-exact; percentile intervals pass it on both grids, basic intervals fail it on both), so $\epsilon{=}0.10$ is the only convention-stable margin and the one claimed.  Both margins exceed the anchor's own reproducibility floor (split-half implied full-$N$ floor $.033$ de-en / $.026$ zh-en; $\epsilon{=}0.10\approx 3\times$ that floor).\footnote{Margin provenance artifacts: r59/E2 (revision-round entry), r64/E-A6 (exploratory scan), r59/r60 (freeze comparison), r64/E-A6a (anchor floors), r64/E-A11 (out-of-sample margin confirmation, newstest2021 en-de).}  Out-of-sample confirmation (response analysis, run for this revision, outside the freeze): on newstest2021 English$\to$German, a grid outside the calibration on both direction and year, anchored by the WMT21 regular human DA (20 systems; 1002 segments; 135 annotators, 20{,}960 segment-level judgments, per-annotator $z$ per the official pipeline; sparse per-system coverage, 555--737 of 1002 segments), the margin to the best single contract reproduces at the claimed width: point $\Delta{=}{-}.011$ ($\tau_b$ $0.505$ AER vs $0.516$ BLEU-13a), with CI90 $[-.063,+.011]$ entirely within $\pm0.10$ (TOST pass) and not within $\pm0.05$ (fail); the equivalence margin is confirmed at $\epsilon{=}0.10$ and not tighter (r64/E-A11).  At the anchor's own noise scale the picture splits: on de-en the TOST fails at $\epsilon{=}.033$ (CI90 lower limit $-.050$, analytic power at the observed point $\approx$ zero) and the margin is not resolvable below $\epsilon{\approx}.05$; on zh-en equivalence holds even at $\epsilon{=}.026$ (power $.14$); and the $\tau_b$ lattice at $n{=}16/15$ systems makes differences below $.017/.019$ unrepresentable in principle (r64/E-A12).

\begin{figure}[htbp]
\centering
\includegraphics[width=0.98\linewidth]{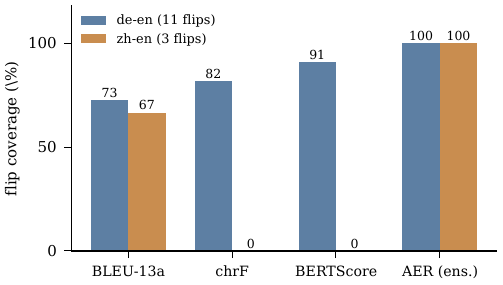}
\caption{Registered-flip coverage on the WMT19 grids (r60 canonical):
the audited ensemble covers every registered flip on both grids; no
single contract does (9--10; McNemar n.s.).}
\label{fig:coverbars}
\end{figure}

The tolerance component \textbf{covers every flip the
falsification found} (Figure~\ref{fig:coverbars}; de-en 11/11, zh-en 3/3); no single contract covers all 14 (9--10; McNemar n.s.).

  The comparison is conservative: the best single is selected post hoc and changes across grids, so an evaluator who cannot know the right contract ex ante gets its level plus full flip coverage.  The framing is declared rather than assumed: contract movement in this paper is a measurement hazard to be bounded, not evidence about which contract is more valid.  The alternative frame (read the disagreement as a validity signal and select the contract that tracks the anchor best) is instantiated in every comparison above as the best-single row, and it is the right strategy exactly where an anchor exists to define \emph{best}. AER is the strategy for grids where no anchor licenses a selection, and it carries the tie declarations that selection does not.  Equal standing for gD and gR in the DocRE ensemble is the same no-assumption default: gR's corrective priority is real (it is the published correction of gD), but it is an annotation-quality judgment external to the contract registry, so the ensembling reports both rather than adjudicating, and a reader holding the normative frame can read the gR rows as the answer.  Scope, explicitly: AER's equivalence and coverage claims are made for scorer/format/language grids without a normative anchor; on the gold-version axis the side-by-side gD/gR rows are the deliverable, and the equal-weight ensemble is the no-assumption example, not a reading to act on (r64/E-A12; cf.~the adoption heuristic above).  The anchor is itself a coordinate: the raw-vs-$z$ anchor choice moves AER by $-0.050$ (de-en) and $-0.038$ (zh-en), with the margins to the best single contract preserved.  Held out (newstest2021; 180 contract-induced flips under the point-level flip definition): AER ensemble covers 73\%, median-of-z 73\%, best single 69\%; the union-of-ties rule (declare tied if any flipping contract does) reaches 89\% but with 72--91\% false-tie rate, meaning a tie is declared while some single contract CI-excludes, a discordance rate rather than an error rate against truth; an AER tie is a failure to reject in both directions, unresolved sign evidence, not an asserted equivalence. This is a coverage--precision trade-off the marginalization choice accepts.  The ensemble's own false-tie cost, at the same all-pairs scope and definition, is $37/56$ ($66.1\%$) on de-en and $22/25$ ($88.0\%$) on zh-en, between BERTScore ($58.7\%/86.4\%$) and BLEU-13a/chrF ($71\%/90\%$; BLEU-13a zh-en reaches $91.4\%$, r64/E-A12b). The missed-flip side is priced symmetrically at the same scope: the ensemble leaves 48 of the 180 held-out flip entries uncovered ($26.7\%$; the median-of-z reading likewise leaves $48/180$, with opposite per-grid directions: $82$ vs.\ $88$ covered on de-en, $50$ vs.\ $44$ on zh-en), so the trade-off is priced on both sides (r64/E-A12b).

Alternative paths not in the comparison are recorded as such.  Threshold-aligned comparison: within a family whose flips concentrate on the threshold coordinate (ten of thirteen registry flips share the fixed-vs-decoded contrast), re-scoring at one common declared threshold is a near-zero-cost competitor that removes the dominant arm; it is contract selection rather than an estimator, and yields neither tie declarations nor a coverage guarantee.  Normative-gold selection: gR is error correction, and equal-weight marginalization gives the defective gold the same voice as the corrected one; the corrected-gold readings are reported side by side (\S\ref{sec:empirical}, gR rows) and the adoption heuristic routes dense-incidence grids to normative-gold adjudication first, but a normative priority inside the ensemble would import an annotation-quality ordering the contract set does not license.  Generic shrinkage: the registered nearest instance is median-of-z, which matches the best single on the judge grid (.717 vs ensemble .700) and ties AER's held-out coverage (73\%/73\%) while surrendering the flip-coverage margin on the declared grids; any data-dependent re-weighting of members is itself a contract and would need its own registration.  Density weighting over a contract posterior has no likelihood under the freeze, and three-to-five members per family cannot support one, unregistered future work.

The tie declarations are robust to the interval method: re-declaring every held-out pair under basic (reverse-percentile) intervals, same bootstrap stream, shifts ensemble coverage by $-1.1$pt (73.3\%$\to$72.2\%) with 96.7\% per-pair verdict agreement (174/180; flip sets near-identical, 4/180 discovered additionally under basic).  The $\epsilon$ bands exceed the anchor's resampling noise (jackknife SD $.010$/$.005$, split-half floor $.033$/$.026$), scoping coverage to the declared grid and member set.  On the DocRE flip registry itself, the same tie machinery covers 3/13 flips (per-family two-contract ensemble) and 5/13 (union-of-ties), against 14/14 on the declared WMT19 grids and 73\% held out: the coverage guarantee is grid-specific, degrading with the density of statistical ties across testbeds.  An ex-ante separator is contract-induced flip incidence, the fraction of system pairs that flip under some declared contract, computable from any frozen grid before adopting AER.  Across the three CI-defined flip registries the ordering is monotone: $8.9\%$ (TACRED, 88/990) and $9.2\%$/$2.9\%$ (WMT19 de-en/zh-en, 11/120 and 3/105) at full coverage (88/88, 14/14), against $57\%$ (12/21) on the DocRE registry at 3/13.  The held-out newstest2021 grids report point-flip incidence (47\%; 180 flips over $381{=}171{+}210$ pairs across the two held-out grids) under a different flip definition and are left outside this comparison pending a CI-defined recount.  Where few pairs are contract-fragile, marginalization absorbs the disagreement; where most are, it cannot.  As a provisional adoption heuristic (three grids, labeled as such, not a calibrated threshold): incidence at or below the high-single digits (all three full-coverage grids sit $\leq 9.2\%$) favors AER-style marginalization; dense grids call for contract selection or normative-gold adjudication instead (\S\ref{sec:boundary}).  Learned metrics enter the contract set only with contamination screening and membership-sensitivity disclosure: MetricX-23 \citep{juraska-etal-2023-metricx} (fine-tuned on WMT DA and MQM ratings through the anchor years, per the released model card) leaves best-single identity unchanged and de-en coverage intact (11/11) while zh-en ensemble coverage drops from 3/3 to 1/3; COMET-22 \citep{rei-etal-2020-comet,rei-etal-2022-comet22} (trained on WMT17--20 DA with an MQM-fine-tuned ensemble component, which subsumes the anchor grids) flips best-single identity in both grids, reaches $\tau_b{=}0.962$ on zh-en (anchor contamination), and collapses de-en ensemble coverage to 5/11; the equivalence reading then fails on zh-en ($\Delta$ CI90 $[-0.21,-0.13]$) not because the ensemble degrades but because its comparator is the contaminated metric.  Both learned metrics therefore sit inside the screening's risk set; the contrast between them is measured (coverage and $\tau_b$ effects), not assumed from training corpora alone.\footnote{Section-artifact map: anchor coordinate r61/E14; held-out coverage and false-tie rates r64/E-A3; basic-interval re-declaration r64/E-A3b; noise floors r64; DocRE registry coverage r64/E-C1; learned-metric contamination r64/E-A4.}

\begin{table}[t]
\centering
\caption{Canonical AER validation on WMT19 (16/15 released systems).}
\label{tab:aerties}
\setlength{\tabcolsep}{4pt}
{\footnotesize
\begin{tabular}{@{}lcc@{}}
\toprule
AER audit ($R{=}2000$, r60) & de-en & zh-en \\
\midrule
point $\tau_b$ (CI95) & \textbf{$.700$} & \textbf{$.752$} \\
best single (post hoc) & $.683$ & $.771$ \\
$\Delta$ vs.\ best & $+.017$ & $-.019$ \\
TOST $\epsilon{=}.05$ / $.10$ & pass / pass & pass / pass \\
flips covered by ties & \textbf{$11/11$} & \textbf{$3/3$} \\
\bottomrule
\end{tabular}
\par\vspace{3pt}\footnotesize
\emph{Notes:} CIs: point $\tau_b$ CI95 $.700\,[.617,.717]$ / $.752\,[.733,.771]$; $\Delta$ CI90 $+.017\,[-.050,+.017]$ / $-.019\,[-.019,+.019]$ (de-en lower limit at the $\epsilon{=}0.05$ band edge; each point estimate lands on a percentile-bootstrap interval endpoint, a known behavior of original-sample points under skewed resample distributions). Margins post-date the 2026-10-03 freeze; $\epsilon{=}0.10$ is the convention-stable choice (\S\ref{sec:aer}); the $\epsilon{=}.05$ pass on de-en is percentile-only and boundary-exact (fails under basic intervals). Ties re-declared at $R{=}2000$; the flip set is identical to $R{=}200$ (pair-for-pair, both grids). Best singles: BLEU-13a (de-en), BERTScore (zh-en); bold marks the audit's headline readings.
}

\end{table}

\subsection{Extension to LLM-as-Judge: Three Judges on a Contract Grid}
\label{sec:judge}

\looseness=-3
The audit extends to LLM-as-judge, whose identity, rubric, format, and language are contract coordinates: three open-weight judges (Qwen-2.5-7B, Llama-3.1-8B, Mistral-7B; three families) score 2{,}000 WMT19 de-en translations from 16 frozen systems under rubric, format, and language contrasts, each cell frozen (r54; anchor: official raw human scores).

Three patterns replicate (Table~\ref{tab:judge}; r54).  Judge-side sampling noise is measured, not assumed: the registered noise ladder repeats each judge five times (majority vote) against its greedy decode (r54/hj2), giving per-judge single-sample segment SDs of $.072$ (Qwen2.5-7B), $.346$ (Llama-3.1-8B), and $.121$ (Mistral-7B-v0.3), with five-vote residuals of $.058$, $.274$, and $.082$: sampling noise is judge-specific and, for two of the three judges, voting reduces it, so judge identity and judge sampling are two separable noise components, and the fixed-facet treatment below is a per-judge operating characterization rather than a claim that judges are deterministic. Mistral-7B shows the largest verbosity sensitivity, and system-level correlation coexists with pair-level unreliability.  Two repairs and an extension sharpen the picture.  (i) The judge-grid ensemble interval $[.70,.70]$ is degenerate by construction: the ensemble reading holds one $z$ per system, so segment resampling cannot move it; reformulating the same reading per segment yields $\tau_b{=}.700$ with 95\% CI $[.683,.750]$ (BCa $[.667,.717]$), and leave-one-system-out jackknife gives SE $.16$ (delete-one convention, $\sqrt{n-1}\approx 3.9\times$ the $.041$ spread of the LOO values themselves; LOO range $[.657,.771]$), the published interval understated uncertainty (r64/E-C2).  A sweep of every r54--r61 statistics file for the broadcast pattern finds no other affected path; the sweep statement ships with the bundle (r64/E-C2).  Recounting sign-unstable pairs under this repaired per-segment ensemble gives 9 (singles $15/14/13$, validated bit-exact under the same bootstrap; the published zero was an artifact of the broadcast layer, where every resample returns identical system means, r64/E-A12).  (ii) The axes extend to prompt format and language: format swaps move arms within ties for both judges ($+.0178$, CI $[-.0006,+.0371]$; $+.0123$, CI $[-.0014,+.0257]$), and the language arm yields a CI-excluding edge under freetext ($-.0099$, CI $[-.0199,-.0002]$) with json tied ($-.0062$) (r64/E-A5).

\begin{table}[t]
\centering
\caption{Judge-grid patterns (three judges, three families; 2{,}000 WMT19 de-en translations, 16 frozen systems. Section~\ref{sec:aer}).}
\label{tab:judge}
\setlength{\tabcolsep}{2.5pt}
{\small
\begin{tabular}{@{}lp{0.56\linewidth}@{}}
\toprule
Pattern & Reading \\
\midrule
(i) Judge leverage & judge swap moves rankings: $\tau_b$ $.683$--$.717$ vs.\ anchor; $z$-gap $.055$ (rubric $.067$, format $.047$); pair-level $\kappa$ $.05$--$.06$ \\
(ii) Judge variance & monotone noise ladders (single-sample SD above five-vote residual); candidate-order flips $.47$--$.63$ of verdicts \\
(iii) AER suppression & unstable pairs $13$--$15 \to 9$ (singles $15/14/13$; per-segment ensemble repair, r64/E-A12); generic shrinkage: median-of-z $.717$ = best single; ensemble $.700$, CI95 $[.683,.750]$ (BCa $[.667,.717]$; r64/E-C2). H-J4 fails, H-J3 met\footnote{H-J3: the ensemble's sign-unstable pair count falls below every single judge's. H-J4: the ensemble's anchor agreement exceeds the best single judge's.} \\
\bottomrule
\end{tabular}}

\end{table}    
\section{Boundary Conditions and Negative Results}
\label{sec:boundary}

Alongside these results, five boundary findings and two failed preregistered hypotheses map where the claims hold and stop (Table~S.13 (supplementary); H-X1 registered a cross-task correlation between arm coefficient and rank volatility (expected $\rho \geq 0.7$, observed $-0.963$, $p = 0.0021$).  A verdict analysis on the frozen per-pair points: the test compares six points from two tasks (three model pairs each); the association is driven entirely by task pooling (leave-one-task-out $\rho = -0.26$ dropping SST-2, $-1.00$ dropping NER), the volatility quantity is scale-mismatched across tasks (median $V$ differs by an order of magnitude between SST-2 and NER), and the third registered task entered only as a demonstration route where leverage is undefined.  The registered test is therefore uninformative as designed: it neither supports nor falsifies the arm$\to$volatility mechanism; we retain the registered failure rather than re-interpret it, and the mechanism question passes to a scale-normalized replication with more tasks.  The findings span SST-2 \citep{socher-etal-2013-sst}, NER \citep{tjong-kim-sang-de-meulder-2003-conll}, XNLI \citep{conneau-etal-2018-xnli}, plus four DocRE-family benchmarks.

\looseness=-3
One pattern stands out: B1 is a failure of naive transfer; the arm needs the disagreement sets of a gold revision, which single-gold benchmarks lack (NER's natural dual gold, strict- vs. relaxed-span readings of the same entities, yields a rescuable fraction far below DocRED's revision density, r36). B3 is a seed-budget bound (sign undetermined at every $n\leq 10$, \S\ref{sec:empirical-llm}).  Two mechanism checks: the X2 gold-swap direction is robust across scoring functions (magnitude varies four-fold, r40); removing entity markers collapses part of the HacRED gap ($-0.006$ to $-0.015$ vs.\ $+0.05$ to $+0.09$): an input-side convention carries part of the reading.  B4 localizes the phenomenon: leverage needs models that err differently \emph{and} score comparably, violated by strong models on easy benchmarks; with B2, B5, H1, H-X1, these scope Propositions 1--4. One inference rule: a flip under frozen predictions shows a single-contract reading cannot be presumed portable without an arm/margin audit; stability within the tested family does not certify a model-intrinsic property.

\section{The Audit Protocol and the EvalLever Toolkit}
\label{sec:audit}

A protocol that audits evaluation must itself be verifiable; this section validates the machinery and releases its implementation.

\paragraph{Fault-injection and positive control.}
\looseness=-3
Five fault classes over 675 scenarios: 100\% detection, zero false positives (Wilson 1.15\%; 330 legal variants), 98.19\% key localization (preregistered $\geq$99\% \emph{failed}).  A seeded oracle: clearing 1\% of intra pairs moves the reading $-0.004$ $[-0.006,-0.002]$, monotone at every level; the sham injection reproduces frozen counts bitwise.

\paragraph{Official-scorer reproduction gates.}
\looseness=-3
Adapters enter only after exactly reproducing official evaluation (digit-matched; supplement).

\paragraph{Human verification with programmatic ground truth.}
\looseness=-3
A stratified 50-violation sample: non-blind author detection 50/50, classification 48/50 (Wilson does not); held-out Chinese-arm violations 20/20; blind second-annotator detection 50/50.  Subcategory agreement on the 21 classification rows is $14/21$ ($\kappa=0.561$, CI $[0.38,0.77]$), decomposed by channel: self-pair $9/9$ and relation-channel $5/5$ agree exactly, and all seven differences are one annotator using a coarser bucket, violation-status agreement is $21/21$.  A documented arbitration of all 36 label differences attributes 30 to label misreads (annotator notes match the definition), 2 to a display convention, 2 to more-precise labeling, and 2 to substantive divergence; the out-of-bounds channel is verified 29/29 against program truth.  Taxonomy percentages remain single-coder descriptions, and the disagreement-arbitration protocol is the documented routing for flagged rows (rev78).
A post-hoc independent-model second coder re-checked the violation-class taxonomy under blinding, a codebook audit, not independent verification: GLM (a frontier LLM not among the audited systems; the same model family as the disclosed writing assistant, in a different task role), given only a written codebook derived from the frozen parser rules and the raw output lines, re-coded a stratified 70-instance sample across the five violation classes; blind Cohen's $\kappa = 0.464$, and the class split carries the reading: the mechanical channels agree near ceiling (out-of-bounds 30/31, self-pair 10/11) while the semantic channels are near floor (parse-fail 0/9, relation-unmapped 2/14, ambiguous 0/5), so the re-coding corroborates the mechanical classes and marks the semantic classes unadjudicable at line level. This is an information asymmetry (the original coder is the frozen parser with the relation registry in hand), not a coder defect.  All 28 disagreements were adjudicated and every one reverted to the original label, with causes decomposed as codebook wording gaps (9), registry-prior absence (12), information unavailable at line level (5), and instance-level judgments (2) (r64/E-A23).

\paragraph{The audit auditing itself.}
\looseness=-3
One ledger reconciliation exposed a parser bug, corrected by re-auditing preserved raw predictions (supplement).

\paragraph{EvalLever.}
\looseness=-3
We release the stack as EvalLever (footnote~1): \texttt{register}, \texttt{aer}, \texttt{audit}, \texttt{seed} (contract set, marginal reading with ties, violation scan, contrast-specific minimum seeds).  Environment: Python $\geq$3.10, standard library plus NumPy; code under MIT, artifacts under CC BY 4.0.

\section{A Preregistered Evaluation Paradigm}
\label{sec:paradigm}

\looseness=-3
\looseness=-4
The components validated above assemble into the operating paradigm this section prescribes.  The workflow (CPU-only throughout; the audit layer itself runs in seconds on frozen predictions, \S\ref{sec:discussion}): \textbf{register} the contract set before running (\texttt{evallever register}); \textbf{report} the marginal reading with ties declared, not ranked, admitting multiple winners (\texttt{evallever aer}); \textbf{gate and prescribe}: violation rate, unverifiable cells, required seed count (\texttt{evallever audit}/\texttt{seed}); amendments logged, negatives retained.  Three evidence tiers (Table~S.3 (supplementary)): T1 preregistered and audit-passed, T2 multi-seed unaudited, T3 single seed; the regress stops at preregistered artifacts and anchors, not another model judge.

\looseness=-1
Against its own standard (Table~S.4 (supplementary)): 147 logged amendments (hypothesis-bearing ones disclosed at point of use), each with a verification record; every negative result retained; four defective launches halted and re-derived.

\section{Discussion}
\label{sec:discussion}

We now turn from evidence to practice: what the audit costs, where it applies, and how its burden is shared.

\paragraph{Deployment cost and decision mapping.}\looseness=-4
For deployed-system practice (model selection, A/B comparisons, release regression), the readings that steer those decisions are exactly the single-contract kind shown here to move; the audit runs on frozen predictions in seconds-to-minutes on CPU, the fourteen claim$\times$gold$\times$seed audits in the stability check of Table~\ref{tab:claims}, each with 10,000-resample bootstrap intervals, complete in 47 seconds single-threaded (Intel i7-10700 CPU, 32 GB RAM) (r56/PS), so contract auditing can enter routine evaluation at checklist cost.  That figure prices the audit layer only: a third party additionally bears prediction generation under $f,\ell$ variants (GPU inference for LLM settings) and needs legal contract alternates to exist (B1); the checklist-cost claim covers the audit step, not the pipeline.  Concretely: a team comparing two retrieval services on frozen logs can run the full register$\to$aer$\to$audit$\to$seed pass before a promotion meeting; an LLM-component owner can route flagged settings to human review using the violation scan and the survivor-selection bound rather than a single metric readout (the scan is a triage flag, not an automated block; none of the seventeen audited settings passes its CI-form gate, \S\ref{sec:empirical-llm}, a non-vacuous bar: the pass ceiling is $3.0$--$4.1\%$ by cell size and the audited cells sit above it (r64/E-A12)); and the evidence tiers give an auditor a vocabulary for stating which readings are confirmatory.  A minimal decision mapping: T1 + gate-pass supports promotion as stated; T2/T3 or gate-flagged readings may inform but not ground a go/no-go, and trigger either human review (violation-driven) or a seed/contract audit (tier-driven) before the next release decision.  The criterion needs released gold-side or scorer-side alternates (\S\ref{sec:boundary}); where only one gold exists the account still disciplines reporting (freeze and state the contract) though leverage itself is then unmeasurable (B1).  \paragraph{Adoption profile by data situation.}
By data situation: released gold alternates with flip incidence at or below the high single digits, marginalize, near-full expected tie coverage (88/88 TACRED, 14/14 WMT19 declared); released alternates with dense incidence (57\% of pairs on DocRE), coverage degrades (3/13), so contract selection or normative-gold adjudication comes first and the audit runs as diagnostic; scorer or format alternates only, format and language arms, near-zero expected for strong models; single gold, freeze-and-state reporting discipline, leverage unmeasurable (B1); LLM outputs, violation-scan triage to human review.  \paragraph{Fairness of adoption costs.}
Contract auditing is resourced work, and the burden is unevenly distributed.  Prediction generation under $f,\ell$ variants takes inference compute; gold-side alternates take annotation campaigns and arbitration labor of the kind \S\ref{sec:audit} shows (two annotators, documented adjudication); and released alternates are less common outside high-resource languages; our own language-axis arms move most exactly where alternates are scarcest (the Chinese HacRED arms; \S\ref{sec:empirical-llm}).  Teams evaluating lower-resource languages can therefore inherit both scarcer alternates and noisier violation taxonomies.  The released toolkit lowers the audit-layer barrier (CPU-only scoring over frozen predictions) but not the alternates-existence barrier; closing that gap is community work (shared alternates and registries) that no single toolkit can substitute.

\section{Design Implications for Neural Architectures}
\label{sec:nn-implications}

This section extracts three architecture-level findings that a neural-network
designer can act on, drawing together results scattered across the audit.

\paragraph{Architecture-to-audit mapping.}  Table~\ref{tab:archmap} condenses
the three findings into the decisions a designer actually faces; the
exposure column is the contract coordinate whose switch moves the reading,
the demarcation principle being that of \S\ref{sec:theory} (officially
released, evaluator-choosable alternates).

\begin{table}[t]\centering\small
\caption{Architecture decisions and their evaluation-contract exposure. Exposure legality differs by row: gold = released-artifact alternates (\S\ref{sec:theory}); threshold and format = evaluator-choice coordinates; hardware/software-stack axes remain unmodeled (Appendix~E (supplementary)).}
\label{tab:archmap}
\begin{tabular}{@{}p{0.30\linewidth}p{0.24\linewidth}p{0.34\linewidth}@{}}
\toprule
Architecture choice & Contract exposure & Audit action \\
\midrule
Sigmoid head vs.\ learned adaptive threshold & Threshold axis; two
calibration families are non-interchangeable ($47$--$61$ $F_1$ points) &
State the operating point per system; never re-score both families at one
common threshold \\
High-recall candidate generation & Gold axis; systematic FP tax under
incomplete gold & Audit the gold-version axis before trusting a leaderboard
gap; prefer released alternates \\
Constrained decoding for LLM structured output & Format$\times$model
violation surface & Treat format as a measurable contract coordinate; route
flagged cells to human review \\
\bottomrule
\end{tabular}
\end{table}

\paragraph{Decoding-head semantics determine evaluation exposure.}
The threshold-axis analysis (Section~\ref{sec:empirical-flip}) reveals that
the audited architectures split into two logit-calibration families:
probability-calibrated (DocuNet, whose sigmoid head peaks near $t{=}0$ with
$F_1$ $0.591$) and ranking-calibrated (the ATLOP family, whose adaptive
threshold head collapses to $F_1$ $0.004$--$.018$ at a fixed sigmoid $0.5$
and rises monotonically toward the grid edge; r64/E-A20).  A designer
choosing between a sigmoid output head and a learned threshold head is
simultaneously choosing the evaluation contract under which the model will
be scored: the two are not interchangeable under a common threshold, and
leaderboard comparisons that re-score both at one operating point
systematically disadvantage the ranking-calibrated family by $47$--$61$
$F_1$ points (r64/E-A19).  This connects to the calibration literature
\citep{guo2017calibration}: the probability-calibrated family is the one
whose confidence scores admit a fixed-threshold reading.

\paragraph{High-recall architectures pay a systematic false-positive tax
under incomplete gold.}
On the gold axis, the false-positive-tax mechanism
(Section~\ref{sec:empirical-flip}; r64/E-A16--A17) shows that architectures
with broader candidate generation pay a systematic penalty when the gold
standard has false negatives, because true predictions on unannotated
pairs are counted as errors.  This is a label-noise phenomenon
\citep{natarajan2013learning,frenay2014classification}: the effective
training signal under incomplete annotation differs from the nominal one,
and models that learn to predict more broadly than the gold annotators
did are penalized for it.  A designer deploying a high-recall architecture
on a benchmark with known incompleteness should audit the gold version
axis before trusting a leaderboard gap.

\paragraph{Violation surfaces are model$\times$format$\times$language
objects.}
The LLM-grid audit (Section~\ref{sec:empirical-llm}) shows that
structural-grounding violations cluster on the interaction of model
identity, output format, and prompt language, not on any single
coordinate.  A designer building a constrained-decoding pipeline
\citep{shin2021constrained} for structured prediction should therefore
treat the format constraint itself as a contract coordinate whose
interaction with the model is measurable, not as a neutral implementation
detail.

\section{Limitations}
\label{sec:limitations}

\looseness=-4
Scope limits: thin pool (\citealp{zhang-etal-2025-global} absent); five LLMs, one decode per cell (the judge grid's five-vote noise ladder excepted, \S\ref{sec:judge}); judge-by-subgroup interactions untested; $n^{\ast}$ from one curve; one verifier (arbitration protocol documented, labor conditions not). DocRE dev-side; released predictions recompute every arm; verdict transfer through retrained stand-ins rests on the fidelity assumption (Lemma~B bounds small-margin sensitivity; the official artifacts degrade through their own entrypoints, Appendix~E).  Ethics: public artifacts only (footnote~1); no personal data; verification: Section~\ref{sec:audit}.

\section{Conclusion}
\label{sec:conclusion}

This paper asked when model rankings move without model change, and answered with a formal account, a registered audit, and a constructive reading.  (i) Capability readings are joint model$\times$contract properties: a point-estimate reversal under a gold switch (frozen; tie-level after), localizing on re-partitioned gold.  This extends the gold-revision re-scoring precedent to the post-2022 generation of published claims, upgraded to tie-level and pair-level registration with an algebraic mechanism: any contract switch whose action is enumerable moves counts only within an explicit set (the gold-diff region on the gold axis; the prediction-set symmetric difference on the threshold axis), so the switched delta is an exact function of that set's composition, identity verified, and flip break points computable; the break-to-gap ratio is thereby a computable fragility criterion for near-tie pairs (r64/E-A16, r64/E-A19).  (ii) Leverage separates families several-fold (descriptive; not monotone).  (iii) A seed prescription follows from the variance--level ordering (dedup-sensitive, imputation-conditional); (iv) violations are orthogonal to leverage (observational).  Freeze, state, and audit the contract: movement CI-excluding, reversal tie-level, findings stated rather than smoothed.  The audited evaluation reading (AER), validated on MT and extended to three LLM judges on one grid (§\ref{sec:judge}), is a first step toward paradigms where the evaluator is itself a language model.

\section*{Supplementary Material}
Appendices A--E (replication details, proofs, statistical procedures, complementary results, and the official-release artifact audit) are provided in the supplementary material file accompanying this submission.

\bibliographystyle{elsarticle-num}
\bibliography{leverage_refs_cas}

\printcredits

\section*{Declaration of competing interest}
The authors declare that they have no known competing financial interests
or personal relationships that could have appeared to influence the work
reported in this paper.

\section*{Declaration of generative AI and AI-assisted technologies in the
manuscript preparation process}
During the preparation of this work the authors used GLM (Z.ai) in order to
assist with language, readability, and copy-editing of the manuscript text.
After using this tool/service, the authors reviewed and edited the content
as needed and take full responsibility for the content of the published
article.

\section*{Data availability}
The complete evidence bundle (code, frozen per-document predictions for
all nine task families, the contract registry, id-level multi-gauge
readings, the preregistration trail with 147 timestamped amendments, and
every cited statistics file) accompanies this submission as supplementary
material and is publicly available at submission:
\url{https://github.com/lutttw/evallever} (v1.0.2, tag-pinned).
Provenance, stated plainly: the freeze timestamp (2026-10-03) and the
147-amendment ledger are self-attested author records; the repository
history was rebuilt before release to remove author emails, so commit
dates are not independent attestations, and per-amendment hashes are
absent for amendments 25--58 (acknowledged in the ledger on 2026-10-10;
the chain resumes at amendment 59); a content-hashed snapshot of this
bundle, ledger included, is deposited at submission under the Zenodo DOI
10.5281/zenodo.24442304, attesting integrity from the deposit date forward.  Bundle
integrity is machine-checkable: the sha256 manifest verifies 321/321 files
(\texttt{python verify\_bundle.py}; \texttt{verify\_counts.py} reconciles
every count quoted here).

\section*{Acknowledgements}
No external funding was received for this work.

\end{document}